\documentclass[letterpaper]{article} 
\usepackage{aaai2026}  
\usepackage{times}  
\usepackage{helvet}  
\usepackage{courier}  
\usepackage[hyphens]{url}  
\usepackage{graphicx} 
\usepackage{natbib}  
\usepackage{caption} 
\usepackage[ruled,vlined,linesnumbered]{algorithm2e}
\SetKwProg{Fn}{Function}{:}{end}
\usepackage{tabularx}
\usepackage{makecell}
\usepackage{multirow}
\usepackage{subcaption}
\usepackage{color}

\usepackage{bm}

\usepackage{pifont}

\newcommand{\argmin}{\mathop{\mathrm{arg\,min}}}

\usepackage{amsthm,amsmath,amssymb}
\usepackage{mathtools} 

\usepackage{algorithmic}
\usepackage{newfloat}
\usepackage{listings}
\DeclareCaptionStyle{ruled}{labelfont=normalfont,labelsep=colon,strut=off} 
\title{Distributional Multi-objective Black-box Optimization for Diffusion-model Inference-time Multi-Target Generation}
\author{
  Kim Yong Tan\textsuperscript{\rm 1}\quad
  Yueming Lyu\textsuperscript{\rm 2,3}\thanks{Corresponding author.}\quad
  Ivor Tsang\textsuperscript{\rm 1,2,3}\quad
  Yew-Soon Ong\textsuperscript{\rm 1,2,3}
}
\affiliations{
  \textsuperscript{\rm 1}College of Computing and Data Science, Nanyang Technological University, Singapore\\
  \textsuperscript{\rm 2}Centre for Frontier AI Research (CFAR), Agency for Science, Technology and Research (A*STAR), Singapore\\
  \textsuperscript{\rm 3}Institute of High Performance Computing (IHPC), Agency for Science, Technology and Research (A*STAR), Singapore\\
  
  \texttt{\{kimyong001, ASYSOng\}@ntu.edu.sg}\\
  \texttt{\{lyu\_yueming, ivor\_tsang\}@cfar.a-star.edu.sg}
}
\nocopyright

\usepackage{bibentry}

\begin{document}

\maketitle

\begin{abstract}

Diffusion models have been successful in learning complex data distributions. This capability has driven their application to high-dimensional multi-objective black-box optimization problem. Existing approaches often employ an external optimization loop, such as an evolutionary algorithm, to the diffusion model. However, these approaches treat the diffusion model as a black-box refiner, which overlooks the internal distribution transition of the diffusion generation process, limiting their efficiency. To address these challenges, we propose the Inference-time Multi-target Generation (IMG) algorithm, which optimizes the diffusion process at inference-time to generate samples that simultaneously satisfy multiple objectives. Specifically, our IMG performs weighted resampling during the diffusion generation process according to the expected aggregated multi-objective values. This weighted resampling strategy ensures the diffusion-generated samples are distributed according to our desired multi-target Boltzmann distribution. We further derive that the multi-target Boltzmann distribution has an interesting log-likelihood interpretation, where it is the optimal solution to the distributional multi-objective optimization problem. We implemented IMG for a multi-objective molecule generation task. Experiments show that IMG, requiring only a single generation pass, achieves a significantly higher hypervolume than baseline optimization algorithms that often require hundreds of diffusion generations. Notably, our algorithm can be viewed as an optimized diffusion process and can be integrated into existing methods to further improve their performance.
\end{abstract}

\section{Introduction}

In multi-objective optimization, we aimed to simultaneously optimize several, often conflicting objective functions~\cite{deb2000fast,moeadde}. Its core difficulty stems from the conflict between objectives, meaning an improvement in one objective may necessitate a compromise in another. Consequently, rather than seeking a single optimal solution, the goal is to discover the Pareto front, which is a set of non-dominated solutions, each representing a unique and optimal balance among the conflicting objectives~\cite{zitzler1999multiobjectiveHV}.

In black-box optimization, the objective function is treated as a black box where only evaluation is allowed and its internal structure is unknown. This is a prominent problem setting for most real-world scenarios, where the objective function is often too complex to be represented in a mathematical form. The key challenge of black-box optimization is that the objective function is non-differentiable, so the problem cannot be solved by efficient, gradient-based algorithms.

In this paper, we address the multi-objective black-box optimization problem. This problem is essential across numerous fields, including drug design~\citep{liu2021drugex,schneuing2024structure,sun2025evolutionary,brown2019guacamol}, engineering design~\citep{kapoor2025surrogate,zhou2024evolutionary}, materials science~\citep{ding2023accelerated,ruan2020surrogate}, and economics~\citep{mcclarren2018uncertainty}, where balancing multiple objectives is critical and the objective functions are treated as black boxes. For example, in drug design, the goal is to optimize drug efficacy while simultaneously maintaining drug selectivity and drug-likeness. These objectives are derived from real-world laboratory experiments or computational simulations, meaning they lack a tractable mathematical representation and are thus treated as black boxes.

Traditional approaches to the multi-objective black-box optimization problem, such as evolutionary algorithms (EAs)~\citep{sharma2022comprehensive,pereira2022review,zhang2007moea}, often suffer from high dimensionality. On the other hand, the diffusion models~\citep{ho2020denoising,dhariwal2021diffusion,esser2024scaling} have been proven very effective in learning complex high-dimensional data distributions. Thus, many recent research~\citep{schneuing2024structure,sun2025evolutionary} has utilized the diffusion models into the optimization algorithm. The powerful generative capabilities of these models allow them to produce realistic data for high-dimensional problems, thereby overcoming the high dimensionality challenges faced by traditional methods.

Much research has been dedicated to this area. One line of research falls in the regime that adjusts the diffusion model itself toward the target distribution, hence allowing it to generate the target data. \citet{lee2024bind,yan2024emodm,annadani2025preference,krishnamoorthy2023diffusion} proposed to first annotate the data with the multi-objective values, then either fine-tune or train the conditional diffusion based on the annotated data. These approaches are known to be time-consuming and data-demanding. On the other hand, \citet{han2023training} train a surrogate model for the black-box multi-objective. The key idea is to approximate the black-box objective with a differentiable surrogate model, so that we can guide the pre-trained diffusion model toward the desired distribution. The key disadvantage is that the surrogate model may not be accurate.

Another line of researches~\citep{schneuing2024structure,sun2025evolutionary} treat the pre-trained diffusion model as a frozen component, acts as a powerful data refiner module in an external optimization loop like an evolutionary algorithm (EA). This approach utilizes a pre-trained diffusion model, thereby avoiding the data-demanding and time-consuming process of training a new one. However, because the diffusion model remains frozen during optimization, the generation model cannot adapt with the distribution shift problem, hence hinders the optimization performance.

To address the challenges mentioned above, we propose a diffusion inference-time multi-target generation (IMG) algorithm. Our IMG algorithm uses weighted resampling techniques to ensure that the samples generated by the diffusion model are distributed according to the desired distribution. Moreover, our IMG supports black-box objectives and does not require a differentiable objective function or surrogate model.

This paper introduces a novel algorithm, Inference-time Multi-target Generation (IMG), for multi-objective black-box optimization using diffusion models. Our primary contributions are:
\begin{itemize}
  \item We first derive the optimal target distribution for the multi-objective optimization problem. We then establish its interesting log-likelihood interpretation, showing it is equivalent to the optimal solution of a minimization problem for a negative log-likelihood objective.
  
  \item Second, we propose the Inference-time Multi-target Generation (IMG) algorithm, which uses a weighted resampling technique to shift the pre-trained diffusion distribution toward our derived optimal multi-target distribution. This inference-time algorithm allow the diffusion model generate a diverse Pareto front in a single diffusion generation pass without additional model training.

  \item Third, as a side-product,  we propose a simple algorithm to generate evenly spaced samples from the uniform distribution on the surface of the positive hyper-sphere by taking advantage of Quasi Monte Carlo.  These samples can be used as a good prior for preference weight vectors. 
  
  \item Finally, we apply IMG to a multi-objective molecule generation task, demonstrating that it achieves significantly higher performance and sample efficiency than strong evolutionary algorithm baselines.
\end{itemize}

\section{Related Works}
\subsection{Diffusion Models}
A diffusion model generates data through a reverse diffusion process, which progressively transforms a simple noise distribution into a target data distribution. The process begins with a sample from a simple prior, typically a standard normal distribution $x_T \sim \mathcal{N}(0,I)$. It then proceeds through a sequence of reverse transitions, where each step denoises the sample $x_t$ into a less noisy version $x_{t-1}$. This transition is governed by a learned probability distribution:
\begin{align}
\label{eq:diffusion}
	x_{t-1} \sim p_\theta(x_{t},t),
\end{align}
which denoises the sample $x_t$ into a less noisy version $x_{t-1}$. The transition probability distribution $p_\theta$ is modeled by a neural network with parameters $\theta$. By iteratively applying these reverse transitions from $t=T$ down to $1$, the diffusion model transforms the initial noise into a final sample $x_0 \sim p_\text{data}(x)$. The transition probability distribution can be written as:
\begin{align}
    p_\theta(x_t,t) = \sqrt{\bar{\alpha}_{t-1}} \hat{x}_{0|t} + \sigma \epsilon_t,
\end{align}
where $\hat{x}_{0|t}$ is the posterior estimation of the clean data given the current state.  For a noise-prediction model, this estimate is given as follows~\citep{chung2022diffusion}:
\begin{align}
    \hat{x}_{0|t} \coloneqq \mathbb{E}[x_0 | x_t] = \frac{1}{\sqrt{\bar{\alpha_t}}} (x_t - \sqrt{1-\bar{\alpha}_t} \epsilon_\theta(x_t,t)),
\end{align}
where $\epsilon_\theta$ is the neural network to predict the noise that was added to the input $x_t$.

\subsection{Optimization using Diffusion Models}

While diffusion models were originally designed for data generation, they can also be repurposed for data optimization. Recent research incorporates pre-trained diffusion models into the optimization loop, effectively treating them as powerful data refiners. In a traditional EA, each optimization step involves two key stages: reproduction and selection. During the reproduction, the process begins by creating new candidates to expand the current population; during the selection, the good candidates (based on fitness function) will be survived and enter next optimization step. 

The reproduction stage is crucial because it generates the new candidates that have the potential to achieve better objective scores than the current population; it directly influencing the EA's overall optimization performance. However, traditional reproduction approaches, such as applying random perturbations (mutation), or combining population members (crossover), can struggle to produce high-quality candidates, especially in high-dimensional spaces. Recent researches use diffusion models to enhance the reproduction stage to generate candidates.

\textbf{DiffSBDD.} One of the pioneering works in this area for molecules generation is DiffSBDD~\citep{schneuing2024structure}. While the paper's main contribution is the generative model itself, the authors also propose an evolutionary algorithm (EA) based on their pre-trained diffusion model. During the EA's optimization step, DiffSBDD systematically adds noise to the current population using the forward diffusion process up to a specific step $\tau$:
\begin{align}
    x_\tau = p(x_\tau | x).
\end{align}
The reverse diffusion process then begins from this noised state, generating new data that serve as candidates for the next generation. 

The authors refer to this strategy as diversify. This approach has two key advantages. First, compared to traditional mutation operators, it better ensures the quality and validity of the new candidates. Second, unlike the vanilla diffusion process that generating samples from scratch (i.e., from pure noise), this method ensures that information is transferred from the parent population to the new candidates. Although the EA algorithm proposed in their work was designed for single-objective tasks, we extend it for multi-objective optimization to serve as one of our baselines.

\textbf{EGD.} Another recent work is EGD~\cite{sun2025evolutionary}, which directly addresses the multi-objective black-box optimization problem for molecule generation. EGD is an EA algorithm also utilize a diffusion model to generate candidates. Similar to the "diversify" strategy in DiffSBDD, EGD also systematically adds noise to the population to obtain $x_\tau$. It then performs crossover on the noised population. This crossover step facilitates information exchange between different members of the population, enabling a more diverse exploration of the search space.

A further distinction lies in its selection mechanism. EGD employs an elitist strategy where both the parent and offspring populations are combined for selection into the next generation. This contrasts with DiffSBDD, where parents are entirely replaced by their offspring. As EGD is a very recent work and its implementation was not open-source at the time of writing, we implemented our own version for the experiments in this paper.

\section{Method}

\subsection{Distributional Multi-Objective Black-box Optimization}

Let $p_\text{base}$ be a any base distribution, and consider a multi-objective optimization problem with $n$ objectives, denoted as $\{ f_1(\cdot),\cdots, f_n(\cdot) \}$.

For each objective $f_k$, our goal is to find a target distribution $q_k(x)$ that minimizes the expected objective value, while remaining close to the original distribution $p_\text{base}$. This is formulated as a KL-regularized distributional optimization problem:
\begin{align}\label{eq:problem}
     q^*_k(\boldsymbol{x} ; \lambda) = \argmin _{q(\boldsymbol{x}) \in \mathcal{P}} \left\{ \mathbb{E}_{q(\boldsymbol{x})} [f_k(\boldsymbol{x})] + \lambda_k \mathrm{KL}(q(\boldsymbol{x})||p_0(\boldsymbol{x}))  \right \},
\end{align}
where $\mathcal{P}$ is the set of all probability distributions family, and $\lambda_k>0$ is a regularization coefficient. The well-known closed-form solution to this problem is an exponentially tilted distribution:
\begin{align}
    q^*_k(\boldsymbol{x}; \lambda) \propto p_\text{base}(\boldsymbol{x}) e^{-\frac{f_k(\boldsymbol{x})}{\lambda_k}}.
\end{align}

To address all $n$ objectives simultaneously, we can model the target distribution as a mixture of these optimal distributions:
Let the mixture of the $k$-th optimal distribution be:
\begin{align}
   q_\text{mix}^*(\boldsymbol{x};\lambda) & = \sum_k \left[ \pi_k q^*_k(\boldsymbol{x};\lambda) \right] \\
   & = \sum_k \left[ \pi_k p_0(\boldsymbol{x}) e^{-\frac{f_j(\boldsymbol{x})}{\lambda_k}} / Z_k \right] \\
   & =p_\text{base}(\boldsymbol{x}) \sum_k \left[ \pi_k/Z_k  e^{-\frac{f_j(\boldsymbol{x})}{\lambda_k}} \right] \\
   & = p_\text{base}(\boldsymbol{x}) \underbrace{\sum_k e^{-\frac{f_k(\boldsymbol{x}) - c_k}{\lambda_k}}}_{W(x;\lambda)} \label{eq:final_q}
\end{align}
where $\pi_k$ are mixture weights satisfying $\sum_k \pi_k = 1$, and $Z_k = \int p_\text{base}(x)\exp (-f_k(x)/\lambda_k) dx$ is the normalization constant for each objective. In the last line, we absorb the constants into a single term $c_k = \lambda_k \log(Z_k / \pi_k)$, rewriting it in a sum of exponential form.

The $W(x;\lambda)$ can be interpreted as a weights factor to transform the base distribution to the target distribution. Hence we can use weighted resampling technique to sample from the $q_\text{mix}^*(\boldsymbol{x};\lambda)$. Specifically, we first sample a batch of candidate from the base distribution, denotes as $X=\{x^1,\cdots,x^M \} \sim p_\text{base}(x)$. Then we can draw a sample from the candidate set according to the categorical distribution:
\begin{align}
\label{eq:sampling_q}
    x^* \sim P(x = x^i) = \frac{W(x;\lambda)}{\sum_{x'\in X} W(x';\lambda)}.
\end{align}
We can also employ greedy strategy which simply pick the sample that has the highest weight.

\subsection{Diffusion-model Inference-time Multi-target Generation (IMG)}

Our goal is to use a diffusion model to generate samples that optimize a set of multiple objectives, denoted as $\{f_1(\cdot),\cdots, f_n(\cdot) \}$. To achieve this, we introduce the Inference-time Multi-target Generation (IMG) algorithm, which adjusts the diffusion model's generative distribution at inference time without requiring model fine-tuning. This algorithm is designed to generate batch of samples to optimize the multiple objectives simultaneously.

The core idea of IMG is to apply a previously derived result to the diffusion process, shifting the pre-trained distribution toward a desired target distribution. Specifically, at each diffusion timestep $t$, we treat the pre-trained reverse transition probability $p_\theta(x_t,t)$ (from Equation~\ref{eq:diffusion}) as the base distribution. We then employ the sampling strategy from Equation~\ref{eq:sampling_q} to draw samples from our target distribution. This results in the new diffusion target transition distribution as follows:
\begin{align}
\label{eq:diffusion_q}
	x_{t-1} &\sim q^*_\text{mix}(x_t,t;\lambda) \\
	&\sim p_\theta(x_t,t) W(x;\lambda).
\end{align}
While this technique allows the model to sample from the target distribution $q^*_\text{mix}(x_t,t;\lambda)$, this distribution is only optimal for a specific preference vector $\lambda$. To generate a diverse set of samples covering various trade-offs, we use batch sampling. For a batch of size $N$, we assign a different preference vector $\lambda^i$ to each instance $i\in [N]$. This allows the diffusion model to generate a diverse batch where each sample represents a unique trade-off, corresponding to a unique target distribution $q^*_\text{mix}(x_t,t;\lambda^i)$. We refer this approach to diffusion multi-target generation.

Specifically, suppose at diffusion step $t$, we have a batch of $N$ noisy instances $\{x^0_t,\cdots, x^N_t \}$. For each instance $x^i_t$, we first generate $M$ candidate samples using the pre-trained model. This creates a buffer containing a total of $B=N \times M$ candidates, constructed as follows:
\begin{align}
	X = [x^1_{t-1},\cdots,x^{NM}_{t-1}], \text{ where } x^{ij}_{t-1} \sim p_\theta(x^i_t,t).
\end{align}

Next, to obtain the final sample $x^i_{t-1}$ for the $i$-th instance, we resample from the entire buffer $X$. This is done by drawing from a categorical distribution where probabilities are determined by the specific preference vector $\lambda^i$. The probability of selecting any candidate for the $i$-th instance is:
\begin{align}
\label{eq:sampling_diffusion_q}
    x_{t-1}^{i} \sim P(x_{t-1} = x_{t-1}^i;\lambda^i) = \frac{W(x_{t-1};\lambda^i)}{\sum_{x_{t-1}^{'}\in X} W(x_{t-1}^{'};\lambda^i)},
\end{align}
where all weight function are with respect to the $i$-th preference vector, and the denominator enumerate over the whole buffer $X$.

\begin{algorithm}[t]
 \caption{Inference-time Multi-Target Generation (IMG)}
 \label{alg:main}

 \KwIn{Pre-trained diffusion model $p$, batch size $N$, resampling size $M$, multi-objective function $f: \mathbb{R}^d \rightarrow \mathbb{R}^n$}
 \KwOut{Set of generated samples $\{\boldsymbol{x}^0_0, \dots, \boldsymbol{x}^N_0\}$}
 \BlankLine

 Initialize preference vector $\boldsymbol{\lambda}^i \in \mathbb{R}^{n} \sim p(\lambda)$ for $i \in [N]$
 
 Sample batch of noise $\boldsymbol{x}^i_T \sim \mathcal{N}(0,I)$ for $i \in [N]$
 
  \For{$t \leftarrow T$ \KwTo $1$}{
    $\tilde{\boldsymbol{x}}^{ij}_{t-1} \sim p_\theta(\boldsymbol{x}^i_{t},t)$ for $i\in[N], j\in[M]$.
    $\boldsymbol{y}^{ij}_{t-1} \leftarrow f(\tilde{\boldsymbol{x}}^{ij}_{t-1})$ for $i\in[N], j\in[M]$.
    
    Collects big batch $B=MN$ of data:\\
     \quad $\boldsymbol{X}_{t-1} \leftarrow \left[ \tilde{\boldsymbol{x}}^1_{t-1}, \dots, \tilde{\boldsymbol{x}}^{B}_{t-1} \right] \in \mathbb{R}^{B \times d}$\\
     \quad $\boldsymbol{Y}_{t-1} \leftarrow \left[ \boldsymbol{y}^1, \dots, \boldsymbol{y}^{B} \right] \in \mathbb{R}^{B \times n}$

   \For{$i \in [N]$}{
   $\boldsymbol{x}^{i}_{t-1}, \boldsymbol{X}_{t-1},\boldsymbol{Y}_{t-1} \leftarrow \mathcal{S}(\boldsymbol{X}_{t-1},\boldsymbol{Y}_{t-1},\boldsymbol{\lambda}^i)$
   }
 }
 
 \KwRet{$\{\boldsymbol{x}^0_0, \dots, \boldsymbol{x}^N_0\}$}

 \BlankLine
 
 \Fn{Preference Sampling Without Replacement
 $\mathcal{S}(\boldsymbol{X},\boldsymbol{Y},\boldsymbol{\lambda})$}{
 
 $b^* = \arg \min_{b \in [B]} \tilde{W}(\boldsymbol{y}^b;\lambda)$
 
 $\boldsymbol{x}^* = \boldsymbol{X}_{b^*}$
 
 Remove the $b^*$-th element from $\boldsymbol{X}$ and $\boldsymbol{Y}$.
     
 \KwRet{$\boldsymbol{x}^*$}
 }
 
\end{algorithm}

\subsection{Pratical Implementation}

\textbf{Pre-compute objective values.} A naive implementation to Equation~\ref{eq:sampling_diffusion_q} is inefficient. Each execution of Equation \ref{eq:sampling_diffusion_q} will enumerate the whole buffer of size $B$, this will consume $B$ objective evaluation. To alleviate this, we first pre-compute the multi-objective values for each candidates in the buffer:
\begin{align}
    y^b = [f_1(x^b), \dots, f_n(x^b)] \in \mathbb{R}^n.
\end{align}
To compute the weight for sampling the $i$-th instance, we then can use the pre-computed objective values:
\begin{align}
    W(x^b;\lambda) \coloneqq \tilde{W}(y^b;\lambda^i) = \sum_k e^{-\frac{y^b_k - c_k}{\lambda^i_k}},
\end{align}
for any $x_b \in X$. We present the overall Algorithm in Algorithm~\ref{alg:main}.

\textbf{Greedy Sampling Without Replacement.} While the probabilistic sampling in Equation~\ref{eq:sampling_diffusion_q} converges to the target distribution with a sufficiently large batch size, the approximation can be inaccurate for smaller batches. To address this, we employ a greedy sampling strategy. Instead of sampling probabilistically, we deterministically pick the best candidate from the buffer for each instance $x^i$. Specifically, we select the candidate that has the largest weight according to the preference vector $\lambda^i$. To ensure diversity within the generated batch, we perform this selection without replacement—once a candidate is chosen, it is removed from the buffer and cannot be selected for any other instance $j\neq i$. Mathematically, the process to select the sample $x^i$ is:
\begin{align}
 b^* &= \arg \min_{b \in [B]} \tilde{W}(\boldsymbol{y}^b;\lambda^i) \\
 x^i &\leftarrow \boldsymbol{X}_{b^*},
\end{align}
where the search for $b^*$ is over all indices of candidates currently available in the buffer.

\textbf{Coefficient.} The calculate for the $W(x;\lambda)$ depends on the coefficient $c_k = \lambda_k \log(Z_k / \pi_k)$ which involves the intractable normalizing term $Z_k$. For practical implementation, we simply set the $c_k$ as the running upper bound throughout the optimization process. That is, the $c_k$ is dynamically set to the worst objective values for all objectives $k\in [n]$.

\textbf{Preference vector.} The instantiation of the preference vector is important as it correspond to the target distribution for each instance in the final samples. The design of the preference vector distribution directly influence the diversity of the resulting samples, especially when the batch size is smaller relative to the objective dimension. We will address the design of the preference vector distribution in following section.

\begin{algorithm}[t]
\caption{Preference Weight Vectors Generation}
\label{alg::GW2}
\KwIn{ Number of the weight vectors $N$, number of  objectives  $n$}
 \BlankLine
   \KwOut{Preference Weight vectors set $\boldsymbol{\Lambda} =[\boldsymbol{\lambda}_1,\cdots, \boldsymbol{\lambda}_N] \in [0,1]^{n \times N}$}
 \BlankLine
\begin{algorithmic}[1]

  \STATE Find the smallest prime number $p$ that satisfies $m\geq 2\times (n-1) + 1$\
  \STATE Construct the generating vector ${\bf{z}} = [1,[N \times \{ 2\cos \frac{{2\pi  \times 1}}{m}\} ],...,[N \times \{ 2\cos \frac{{2\pi  \times (n - 2)}}{m}\} ]] \in {Z^{n - 1}}$\ \\

  \FOR {$j$ =1 to $N$}
  \STATE $\Omega_j = \left\{ {\frac{{j \cdot {\bf{z}}}}{N}} \right\}$\    // assign for $j^{th}$-row of matrix $\Omega$
  \ENDFOR\\

  \STATE Uniformly Sample  $\Delta \in [0,1]^{n-1}$ from unit cube. 
  \STATE Set  $ \Omega =  \{ \Omega + \Delta  \} $, $r = \left\lceil {\frac{{n - 1}}{2}} \right\rceil $,  and $k = \left\lfloor {n/2} \right\rfloor $ 
  
 \STATE Set $\Theta = \Omega_{1:r,:} $, $X = \Omega_{r+1:n-1,:}$
 
 \STATE Set $\Theta = \frac{\pi }{2} \times \Theta $
 \IF {mod(n,2) = 1}
 \STATE $Y_0=\bf{0} ^{1 \times N}$ \
 \STATE $Y_k=\bf{1} ^{1 \times N}$ \
 \FOR {i=k-1:1}

 \STATE  $Y_i = Y_{i+1} X_{i,:}^{1/i} $ // elementwise product \

 \ENDFOR
 \FOR {i=1:k}
 \STATE  $\boldsymbol{\Lambda}_{2i-1} = \sqrt {{Y_i} - {Y_{i - 1}}} \cos {\Theta _i}$ \
    \STATE  $\boldsymbol{\Lambda}_{2i} = \sqrt {{Y_i} - {Y_{i - 1}}} \sin {\Theta _i}$ \
 \ENDFOR
 \ELSE
 \STATE $Y_{k+1}=\bf{1} ^ {1 \times N}$ \
 \FOR {i=k:1}
 \STATE  $Y_i = Y_{i+1}   X_{i,:}^{2/(2i-1)} $  // elementwise product\
 \ENDFOR
 \STATE $\boldsymbol{\Lambda}_1=\sqrt{Y_1}$\
 \FOR {i=1:k}
 \STATE  $\boldsymbol{\Lambda}_{2i} = \sqrt {{Y_{i+1}} - {Y_{i }}} \cos {\Theta _i}$ \
    \STATE  $\boldsymbol{\Lambda}_{2i+1} = \sqrt {{Y_{i+1}} - {Y_{i }}} \sin {\Theta _i}$ \
 \ENDFOR
 \ENDIF
  \RETURN  $\boldsymbol{\Lambda}$
\end{algorithmic}
\end{algorithm}

\subsection{A Negative Log-likelihood Perspective}

The mixture distribution in Equation \eqref{eq:final_q} can be interpreted as the solution to a different, single-objective optimization problem. Inspired by~\citep{lin2024smooth}, we first define a negative log-likelihood function of a Boltzmann distribution w.r.t. aggregation of all objectives given the preference vector $\lambda$:
\begin{align}
\label{eq:nll}
  \mathcal{L}(\boldsymbol{x} ; \boldsymbol{\lambda}) \coloneqq -\log(\sum_k e^{-\frac{f_k(\boldsymbol{x}) - c_k}{\lambda_k}}).
\end{align}
Now, consider optimizing this negative log-likelihood $\mathcal{L}$ in a distribution KL-regularized optimization problem:
\begin{align}
\label{eq:nll_problem}
     q^*_{\mathcal{L}}(\boldsymbol{x};\lambda) = \argmin _{q(\boldsymbol{x}) \in \mathcal{P}} \left\{ \mathbb{E}_{q(\boldsymbol{x})} [\mathcal{L}(\boldsymbol{x} ; \boldsymbol{\lambda} )] + \beta \mathrm{KL}(q(\boldsymbol{x})||p_\text{base}(\boldsymbol{x}))  \right \}
\end{align}
The optimal solution for this problem is:
\begin{align}
q^*_{\mathcal{L}}(\boldsymbol{x}) &\propto p_\text{base}(x) e^{-\frac{\mathcal{L}(x;\lambda)}{\beta}} \\
&\propto p_\text{base}(\boldsymbol{x}) \left[ \sum_k e^{-\frac{f_k(\boldsymbol{x}) - c_k}{\lambda_k}} \right]^{\frac{1}{\beta}}.
\end{align}

Notably, when the temperature parameter $\beta=1$, we find that $q_\text{mix}^*(\boldsymbol{x}) \propto q^*_{\mathcal{L}}(\boldsymbol{x})$. This provides a key insight: sampling from the mixture of optimal distributions is equivalent to minimizing a single, well-defined negative log-likelihood objective.

\begin{table*}[t!]

  \centering
  \caption{Performance Comparison of IMG and Baseline Methods. The table reports hypervolume (mean $\pm$ std), number of Pareto front solutions, and total run time at each objective evaluation. Statistics are derived from three independent runs.}
  \label{tab:result}
  \renewcommand{\arraystretch}{1.1}
  \begin{tabular}{|c|c|c|c|c|}
    \hline
    \makecell[c]{\textbf{Number of} \\ \textbf{Objective Evaluations}} & \textbf{Algorithm} & \makecell[c]{\textbf{Hypervolume} \\ \textbf{(mean $\pm$ std)}} & \makecell[c]{\textbf{Number of} \\ \textbf{Pareto Front}} & \textbf{Run Time} \\
    \hline
    \multirow{4}{*}{25.6k} & IMG (ours) & \makecell{$\textbf{0.5732} (\pm 0.0387)$} & \makecell{12.33} & \textbf{1h 12m} \\ \cline{2-5}
     & EGD & \makecell{$0.5379 (\pm 0.0301)$} & \makecell{17.00} & 2h 34m \\ \cline{2-5}
     & DiffSBDD-EA (Mean) & \makecell{$0.5366 (\pm 0.0374)$} & \makecell{9.00} & 2h 39m \\ \cline{2-5}
     & DiffSBDD-EA (Spea2) & \makecell{$0.5149 (\pm 0.0375)$} & \makecell{18.00} & 2h 41m \\
    \hline
    \multirow{4}{*}{51.2k} & IMG (ours) & \makecell{$\textbf{0.6450} (\pm 0.0964)$} & \makecell{12.00} & \textbf{1h 59m} \\ \cline{2-5}
     & EGD & \makecell{$0.5747 (\pm 0.0480)$} & \makecell{18.00} & 5h 8m \\ \cline{2-5}
     & DiffSBDD-EA (Mean) & \makecell{$0.5619 (\pm 0.0501)$} & \makecell{8.00} & 5h 18m \\ \cline{2-5}
     & DiffSBDD-EA (Spea2) & \makecell{$0.5253 (\pm 0.0623)$} & \makecell{10.00} & 5h 22m \\
    \hline
    \multirow{4}{*}{102.4k} & IMG (ours) & \makecell{$\textbf{0.6972} (\pm 0.0394)$} & \makecell{7.67} & \textbf{3h 42m} \\ \cline{2-5}
     & EGD & \makecell{$0.5732 (\pm 0.0396)$} & \makecell{19.00} & 10h 13m \\ \cline{2-5}
     & DiffSBDD-EA (Mean) & \makecell{$0.5824 (\pm 0.0373)$} & \makecell{3.00} & 10h 37m \\ \cline{2-5}
     & DiffSBDD-EA (Spea2) & \makecell{$0.5515 (\pm 0.0318)$} & \makecell{19.00} & 10h 44m \\
    \hline
    204.8k & IMG (ours) & \makecell{$\textbf{0.7413} (\pm 0.0119)$} & \makecell{13.00} & \textbf{7h 24m} \\
    \hline
    \makecell[c]{32k (for EGD) + \\ 51.2k (for IMG)} & EGD+IMG (ours) & \makecell{$\textbf{0.7447} (\pm 0.0496)$} & \makecell{13.67} & \textbf{5h 11m }\\
    \hline
  \end{tabular}
\end{table*}

\section{Quasi-Monte Carlo sampling for $p(\boldsymbol{\lambda})$}


When the user-specific preference distribution $p(\boldsymbol{\lambda})$ is not given, we can employ the uniform distribution on the surface of the positive hyper-sphere as a good prior.   To sample from $p(\boldsymbol{\lambda})$ more evenly, we propose a simple algorithm (Alg.~\ref{alg::GW2}) by taking advantage of Quasi Monte Carlo sampling.  

\begin{figure}[h]
  \centering

  \begin{subfigure}[t]{0.45\linewidth}
    \centering
    \includegraphics[width=\linewidth]{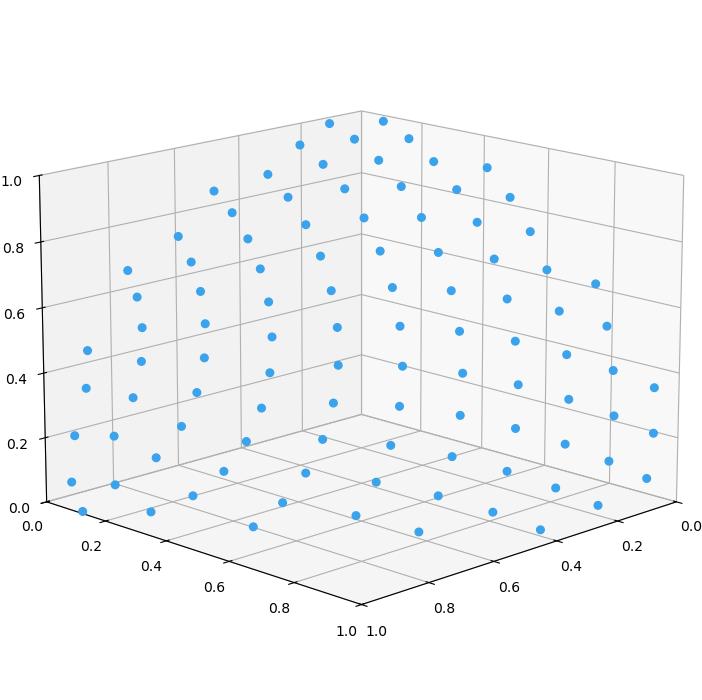}
    \caption{100 samples generated by Alg~\ref{alg::GW2}}
    \label{figw1a}
  \end{subfigure}
  \hfill
  \begin{subfigure}[t]{0.45\linewidth}
    \centering
    \includegraphics[width=\linewidth]{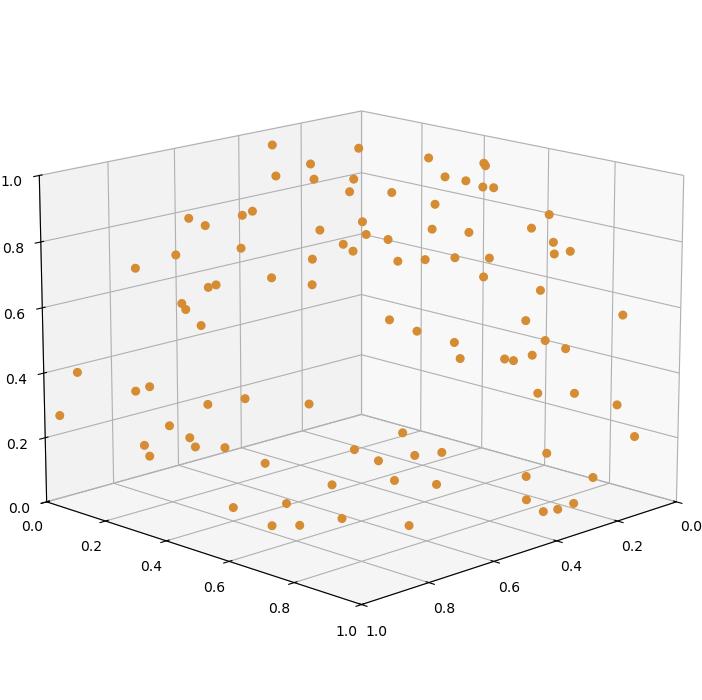}
    \caption{100 samples generated by \cite{Spoints}}
    \label{figw1b}
  \end{subfigure}

  \caption{Demonstration of 100 preference vectors in 3-objective space generated by Alg~\ref{alg::GW2} and \cite{Spoints}.}
  \label{S100}
\end{figure}

 The idea of Alg~\ref{alg::GW2} is that first generate $N$ Quasi Monte Carlo points $\Omega$ in the $n-1$ dimensional unit cube (i.e., $\Omega \in [0,1]^{ (n-1)\times N}$) via lattice rule~\cite{hua2012applications}, then attain the $ \Theta$ and $X$ by projecting $\Omega$ into the complementary $\left\lceil {\frac{{n - 1}}{2}} \right\rceil$ and $n-1-\left\lceil {\frac{{n - 1}}{2}} \right\rceil$ dimensional subspace respectively. Finally, with the contrsuted QMC points $ \Theta$ and $X$ in the unit cube,  Alg.~\ref{alg::GW2} employs the method in \cite{Spoints} to achieve the preference weight vectors $\boldsymbol{\Lambda}$. Note that different from \cite{Spoints}, the Alg \ref{alg::GW2} only needs the preference weight vectors in the first quadrant, thus, the angles $\Theta$ set to $[0,\frac{\pi }{2}]$ in this work.
In Alg.~\ref{alg::GW2}, $\{\cdot\}$ denotes the operation that takes the positive fraction part of the input number element-wisely, i.e., $\{x\}:= x - \lfloor {x} \rfloor$.  And $[\cdot]$ denotes rounding the number to the closest integer.  A Demonstration of gerneted samples from Alg.~\ref{alg::GW2} can be found in Fig.~\ref{S100}.

\section{Experiment}
\begin{figure}[ht!]
    \centering
    \includegraphics[width=\linewidth]{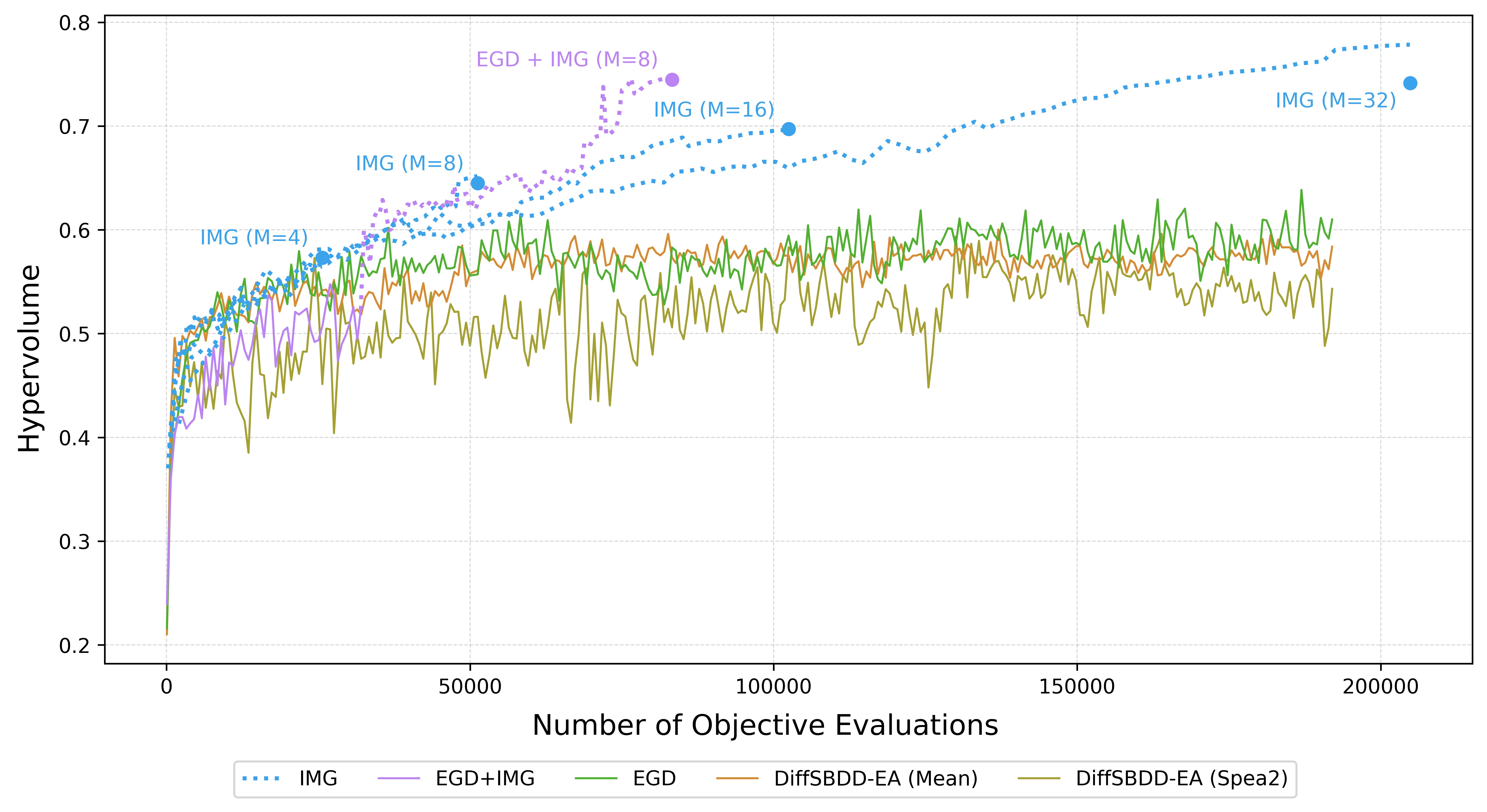}
    \caption{Comparison of hypervolume progression against the number of objective evaluations for all algorithms. Results are averaged over three runs, with dotted lines showing the hypervolume of IMG's intermediate noisy states and solid lines showing the baselines' running population. Experiment demonstrates IMG's superior performance and scalability with increasing resampling size $M$.}
    \label{fig:main}
\end{figure}

\textbf{Pre-trained Model.} We use DiffSBDD\footnote{Available at: \url{https://github.com/arneschneuing/DiffSBDD}}~\citep{sun2025evolutionary} as the pre-trained molecules generative model. It is a conditional diffusion model that can generate realistic 3D molecules conditioned on a target protein's binding pocket. We use its \texttt{crossdocked\_fullatom\_cond} model which was trained on the CrossDocked dataset of 100,000 protein-ligand complexes.

\textbf{Multi-Objectives.} We formulate the drug design task as a multi-objective black-box optimization problem with three objectives. Following the experimental setup in DiffSBDD, we focus on generating oncology inhibitors targeting a specific phosphoprotein (PDB ID: \texttt{5ndu}). The first objective is to maximize binding affinity to this target, which we estimate using the \underline{Vina score}. The second objective is to ensure high synthesizability, measured by the \underline{SA score};and the third is to promote drug-likeness, measured by the \underline{QED value}. 

\textbf{Molecules Generation.} Instead of starting the diffusion generation process from a random Gaussian noise, $\boldsymbol{x}_T \sim \mathcal{N}(0,I)$, we follow the DiffSBDD approach and start from a noisy reference molecule. Specifically, given a reference molecules $\boldsymbol{x}_\text{ref}$, we add noise to obtain $\boldsymbol{x}_\tau \sim p(\boldsymbol{x}_\tau|\boldsymbol{x}_\text{ref})$, where the $p(\boldsymbol{x}_\tau|\boldsymbol{x}_\text{ref})$ is the forward diffusion process as is described in~\citep{schneuing2024structure} Equation 1. DiffSBDD refer to this as a diversify strategy, which allows the model to utilize information from existing reference molecules. Although our Algorithm~\ref{alg:main} describes a general diffusion process starting from Gaussian noise, it can be easily adapted for DiffSBDD by starting the process at step $t=\tau$ with the noised molecule $\boldsymbol{x}_\tau$. For molecule generation, we follow the DiffSBDD implementation which condition the generation on the \texttt{5ndu} target protein, and use the \texttt{8V2} reference molecules. 

\textbf{Objective Evaluation.} We use the source code provided by DiffSBDD to compute the objective values. We flip the sign of all objectives so that lower values are better, aligning with our formulation as a minimization problem. We then normalize these values to the range of $[-1,0]$ to ensure numerical stability and to simplify the subsequent hypervolume calculation.\footnote{While the normalized Vina score may fall below $-1$ due to its undefined lower bound, this does not affect the hypervolume calculation, as the upper bound remains fixed at $0$.}

\textbf{Parameters.} We use batch size $N=64$ and resampling batch size $M \in \{4,8,16\}$. We follow the default DiffSBDD setting with $\tau=100$ diversify steps. In total, this requires $N \times M \times \tau$ objective function evaluations per run.

\textbf{Baselines Details.} We compare our algorithm against two baseline methods: EGD~\citep{sun2025evolutionary} and DiffSBDD-EA~\citep{schneuing2024structure}. Since DiffSBDD-EA is designed for single-objective optimization, we adapt it for our task by aggregating the multiple objectives into a single fitness function. We test two aggregation strategies. The first is to simply take the mean of all $n$ objectives; we refer to this variant as DiffSBDD-EA (Mean). For the second strategy, we follow EGD to use the more advanced SPEA2~\citep{zitzler2001spea2} fitness function, which we refer to as DiffSBDD-EA (SPEA2). For all baselines, we set the population size to $64$, and the total number of evolutionary steps to $3000$. Therefore, a full run for each baseline consumes a total of $64 \times 3000 = 192k$ objective evaluations. All other hyperparameters were kept at their default values as specified in the original publications.

\textbf{IMG Combined with Baselines.} Our IMG algorithm can be readily integrated with existing baseline methods to further improve their performance. While baseline approaches treat the diffusion model as a frozen refiner within an EA loop, IMG directly optimizes the generation distribution itself. To demonstrate this synergy, we created a hybrid method: we first ran an EA algorithm for $500$ steps and then used its final population as the starting state for an IMG generation with a resampling size of $M=8$.

\textbf{Performance Evaluation.} In a multi-objective optimization problem, the objectives conflict with each other. So rather than finding a single optimal point, our goal is to obtain a set of optimal solutions, known as a Pareto front, where each solution represents a different trade-off between the conflicting objectives. And we want to measure the performance of the given set of solutions produced by our algorithm and baselines. One of the most common metric is hypervolume (HV). It measures the volume bounded by the Pareto front to the reference point. Since our objective is upper bounded by $0$, we assume origin reference point $\boldsymbol{0}\in\mathbb{R}^n$. Specifically, given a set of solutions $\mathcal{X} = \{x_1,x_2,\cdots, x_N \}$, the HV is calculate as:
\begin{align}
	HV(\mathcal{X}) = \Lambda \left( \bigcup_{x \in \mathcal{X}} [f_1(x), 0] \times [f_2(x), 0] \times \cdots [f_n(x), 0] \right),
\end{align}
where the $\times$ denotes the Cartesian product, and the $\Lambda$ is the Lebesgue measure.

\textbf{Experiment Setup.} we ran each algorithm three times independently with different seeds, optimizing for the same multi-objective problem. We then report the average hypervolume across these three runs against the number of objective evaluations in Figure~\ref{fig:main} and Table~\ref{tab:result}. A single "objective evaluation" on the x-axis corresponds to the complete assessment of one sample against all n objective functions.

For our IMG algorithm, the hypervolume is calculated at each reverse diffusion step using the intermediate noisy states (i.e., the hypervolume of $\{ x^1_{t-1}, \cdots, x^N_{t-1} \}$ at Algorithm~\ref{alg:main} Line 9). Since these are not the final generated samples, we represent their performance with a dotted line in the figure. For the baseline methods, we report the hypervolume of their running population at each optimization step. 

\textbf{Experiment Result.} As shown in Figure~\ref{fig:main}, our algorithm with $M=4$ achieves a hypervolume comparable to the baseline methods for the same number of objective evaluations. For $M=\{8,16,32\}$, our IMG algorithm achieves a significantly higher hypervolume than the baselines.

Notably, the experiments highlight our algorithm's excellent scalability. Its performance continues to improve with more objective evaluations, in sharp contrast to the baseline methods, whose performance gain almost flattens after 50k objective evaluations. This is likely because the baseline EA-based algorithms treat the pre-trained diffusion model as a frozen refiner, constraining the optimization distribution to the model's original distribution.

Furthermore, as shown in Figure~\ref{fig:main}, executing IMG on the converged EGD population leads to significant further performance gains. This demonstrates that our IMG algorithm can overcome the distributional constraints of the frozen diffusion model that limit baseline methods. 



\section{Conclusion}

We introduced the Inference-time Multi-target Generation (IMG) algorithm, a novel method for multi-objective black-box optimization that steers pre-trained diffusion models toward a diverse set of optimal solutions. By employing a principled weighted resampling technique during the reverse diffusion process, IMG efficiently generates solutions approximating the Pareto front within a single inference pass, eliminating the need for model retraining. This approach is grounded in a theoretical framework that defines an optimal multi-target Boltzmann distribution, directly addressing the distributional shift problem that limits existing methods.

Our experiments on a multi-objective molecule generation task demonstrate IMG's superior performance and efficiency. The algorithm achieves a significantly higher hypervolume compared to strong evolutionary algorithm baselines, and its performance scales effectively with an increased computational budget. Furthermore, we showed that IMG is a flexible module that can be integrated into iterative optimization framework.

\bibliography{aaai2026}


\newpage

\section{Appendix}

\subsection{IMG's Generated Molecules}

\begin{figure*}[ht]
    \centering
    \includegraphics[width=0.3\textwidth]{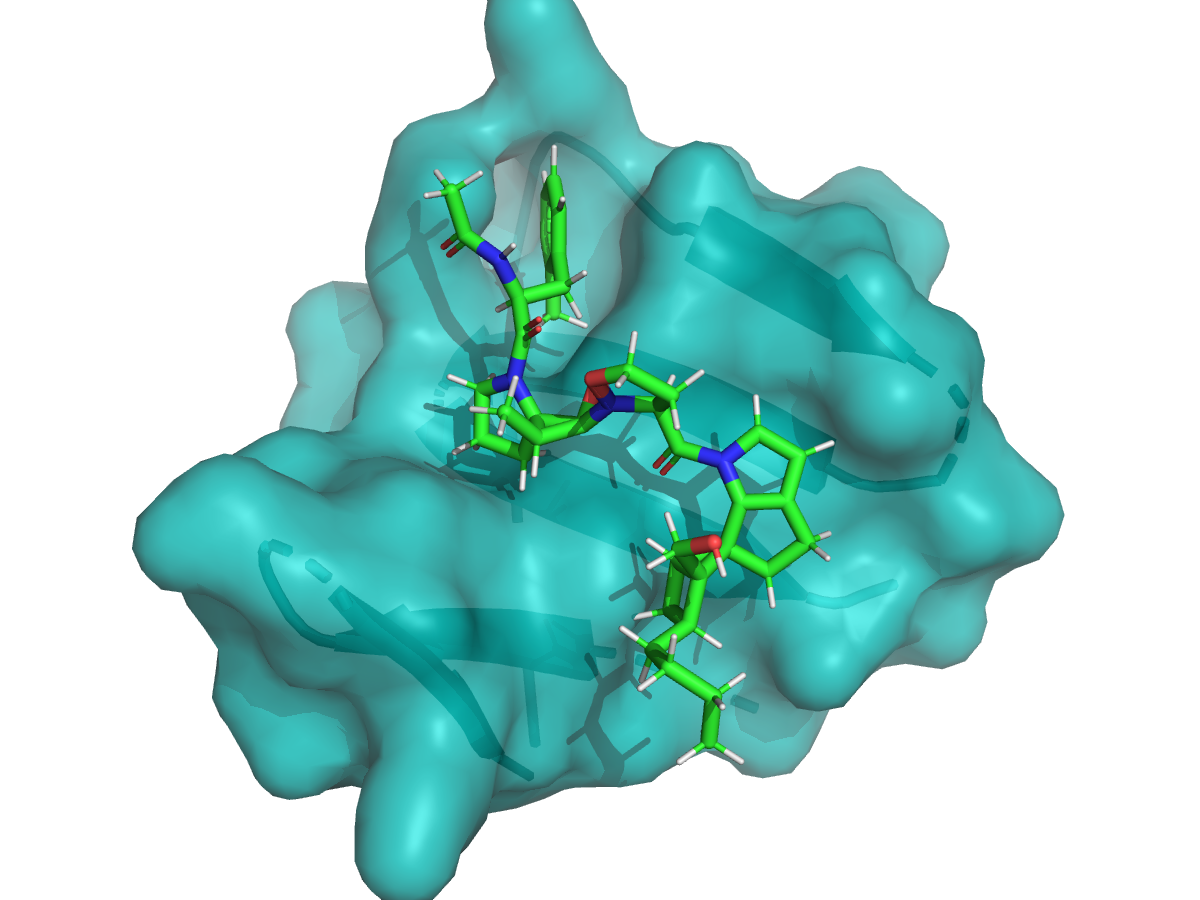} \hfill
    \includegraphics[width=0.3\textwidth]{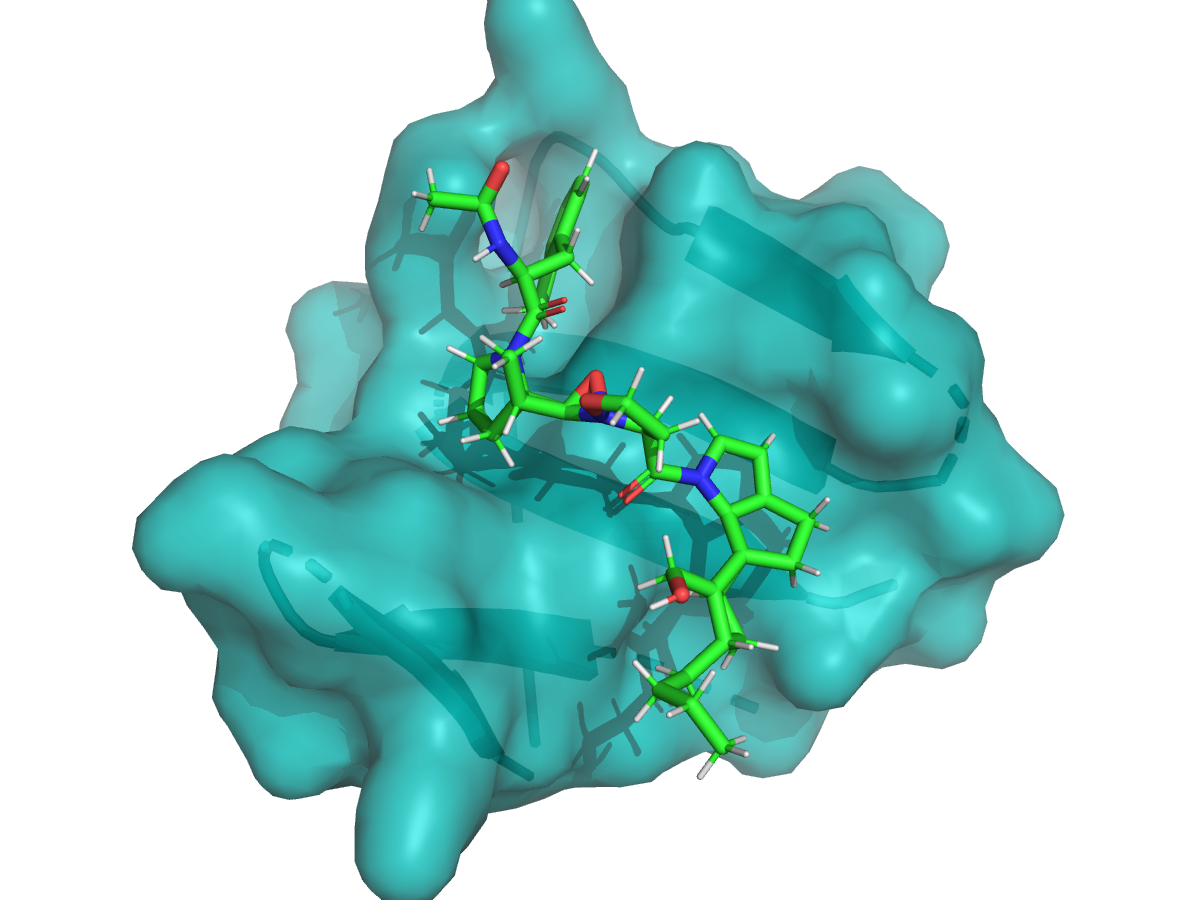} \hfill
    \includegraphics[width=0.3\textwidth]{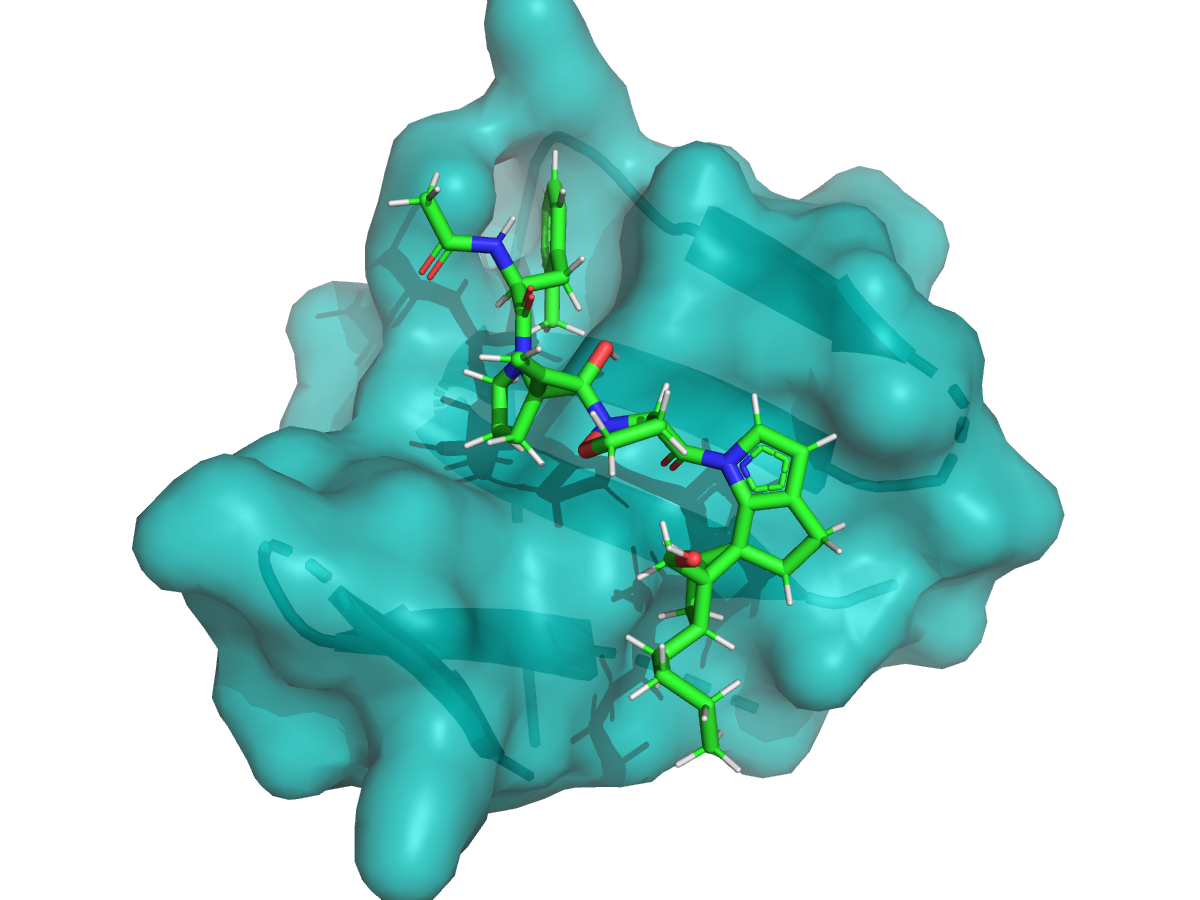}
    
    \includegraphics[width=0.3\textwidth]{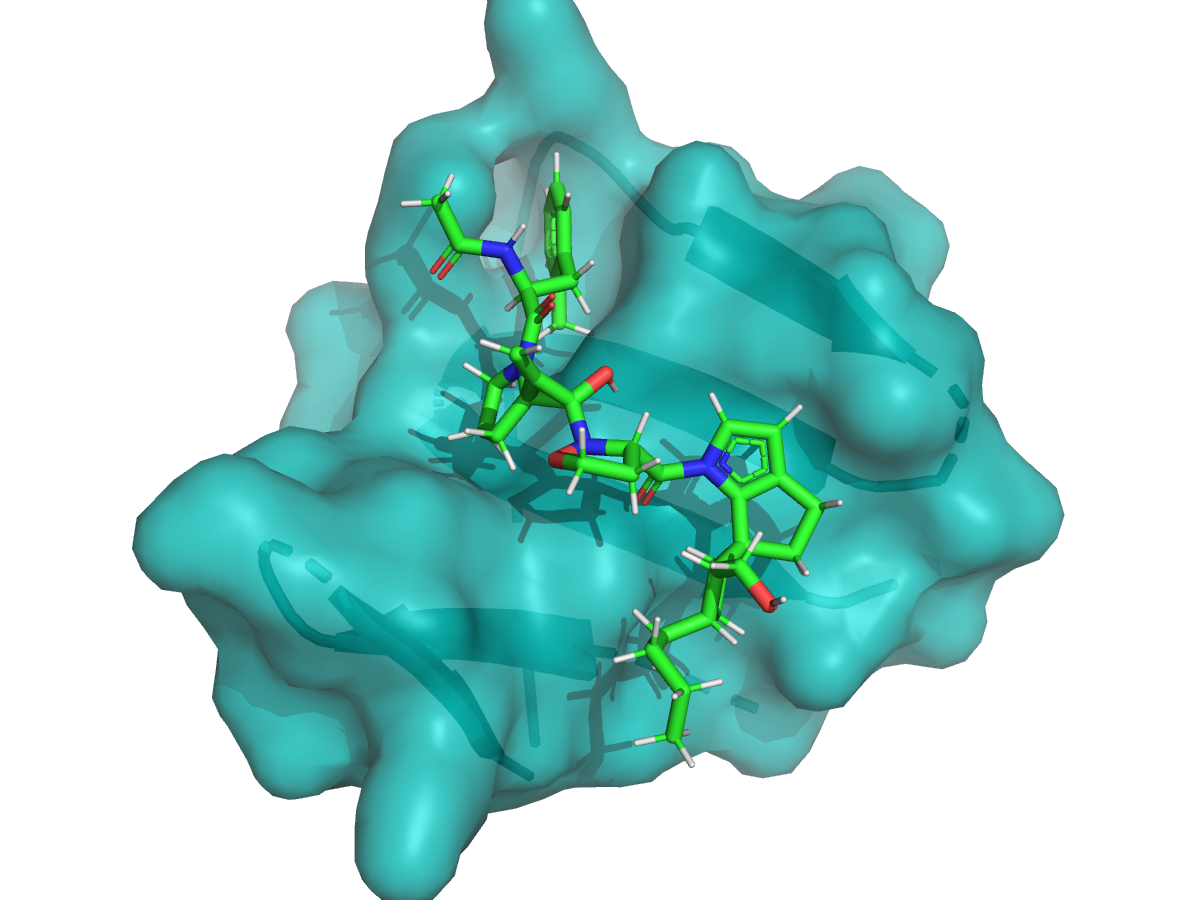} \hfill
    \includegraphics[width=0.3\textwidth]{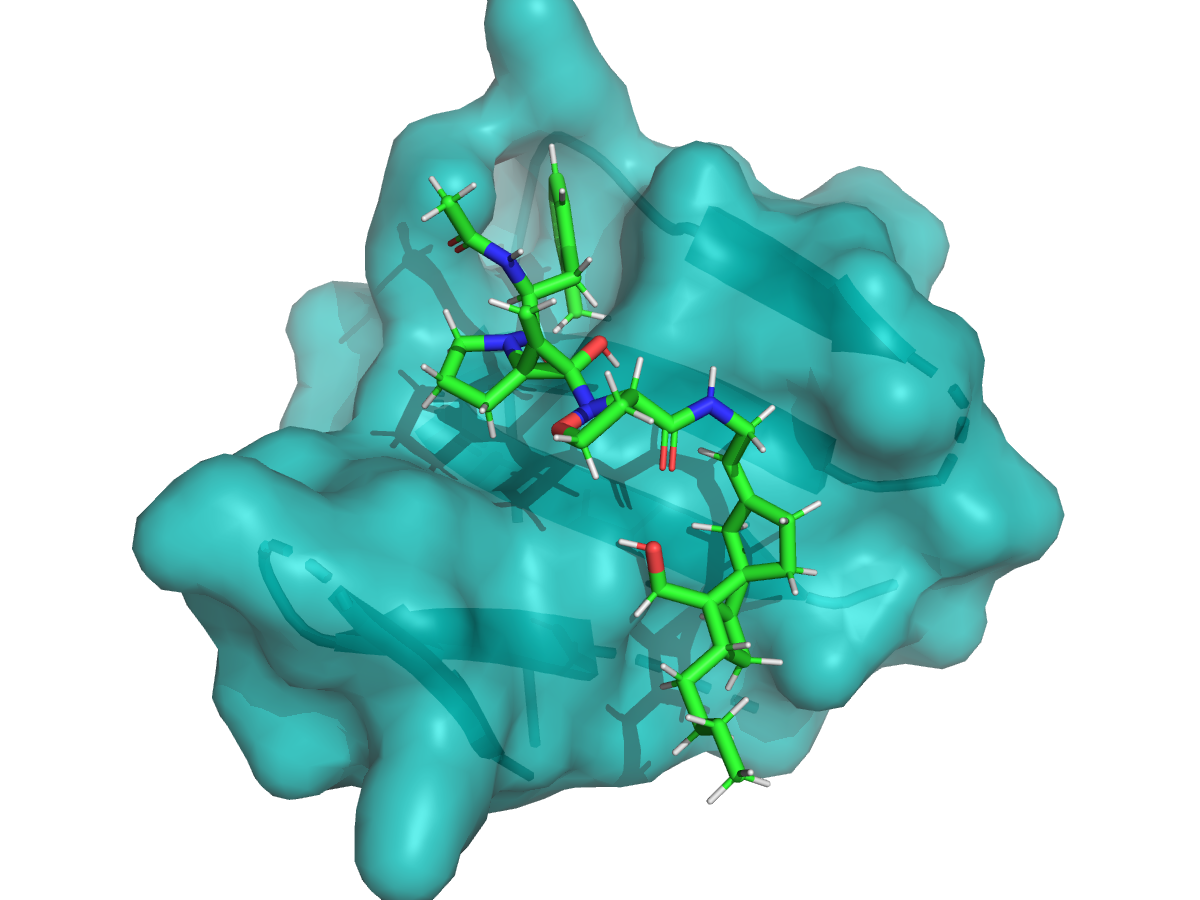} \hfill
    \includegraphics[width=0.3\textwidth]{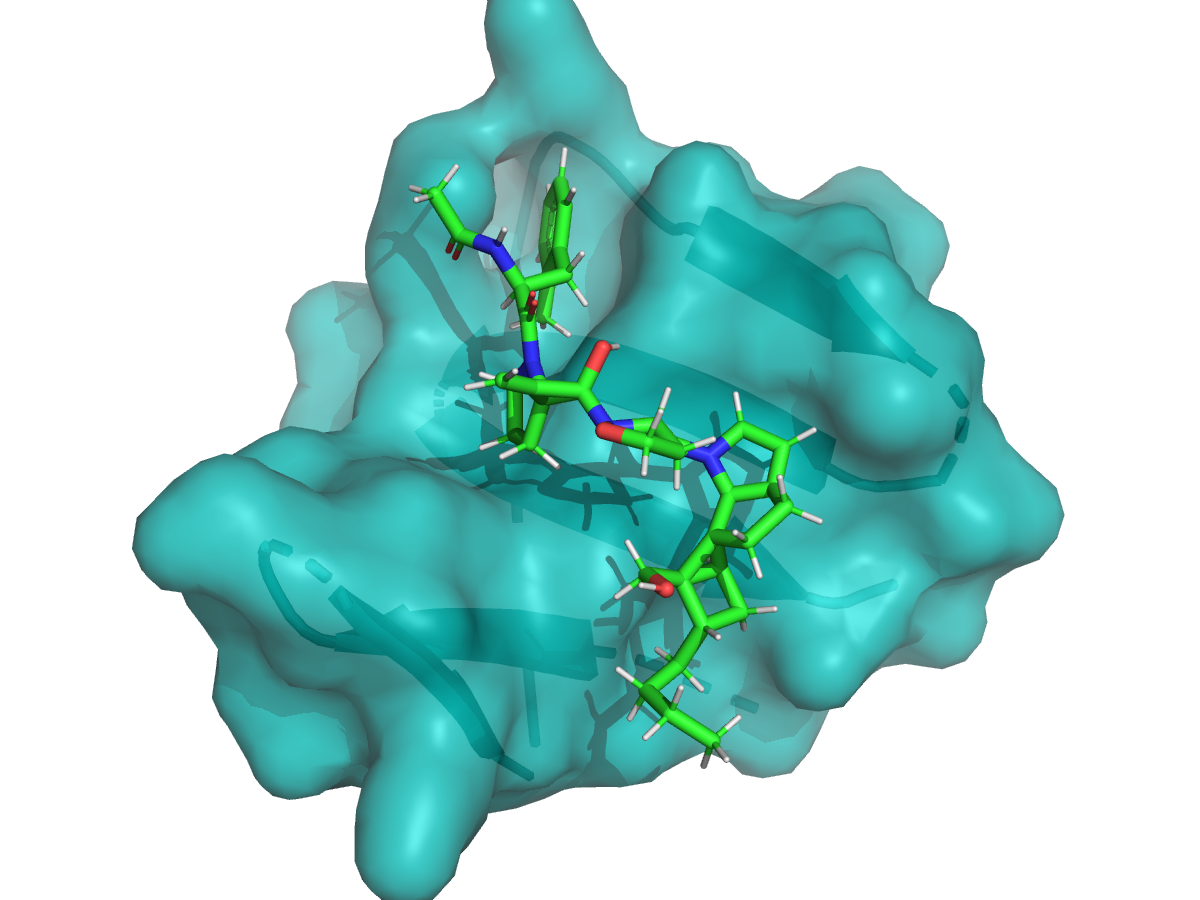}
    
    \includegraphics[width=0.3\textwidth]{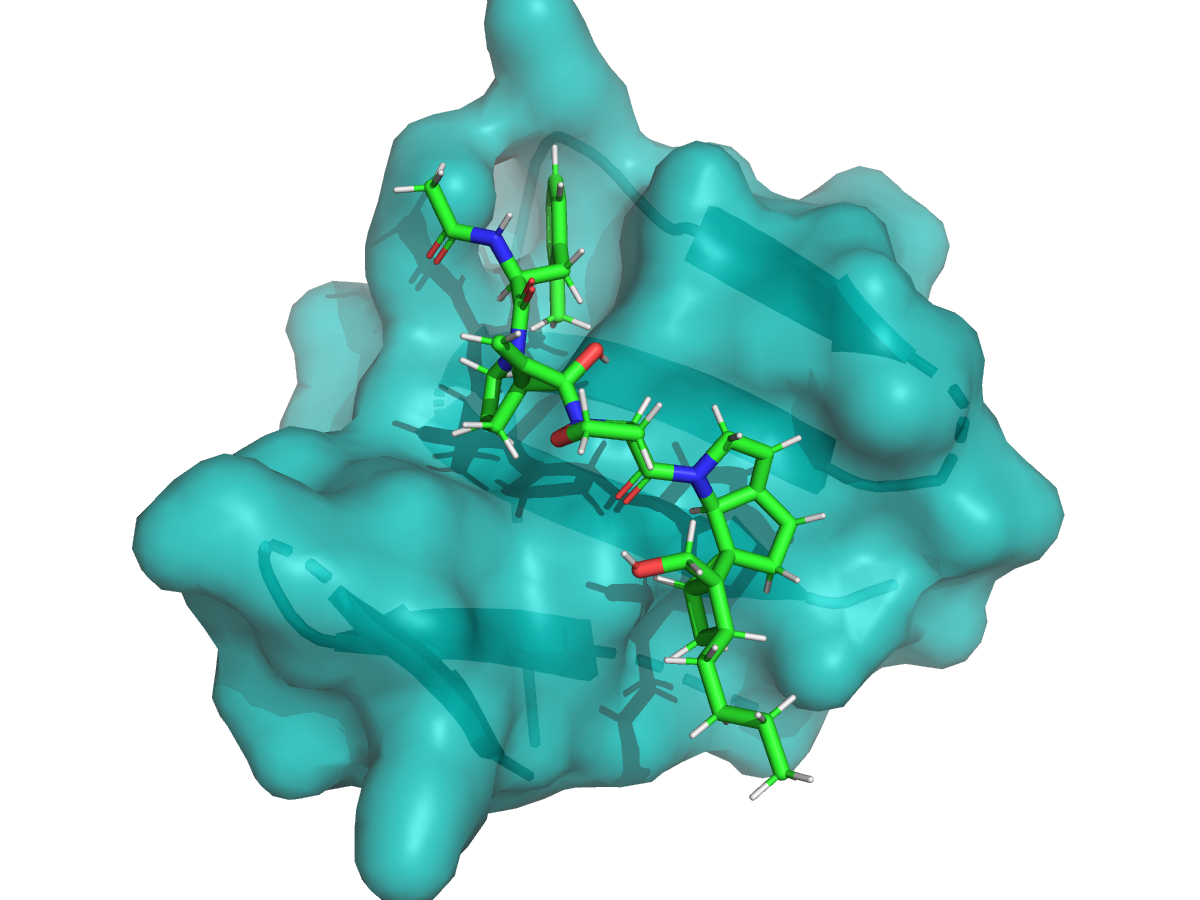} \hfill
    \includegraphics[width=0.3\textwidth]{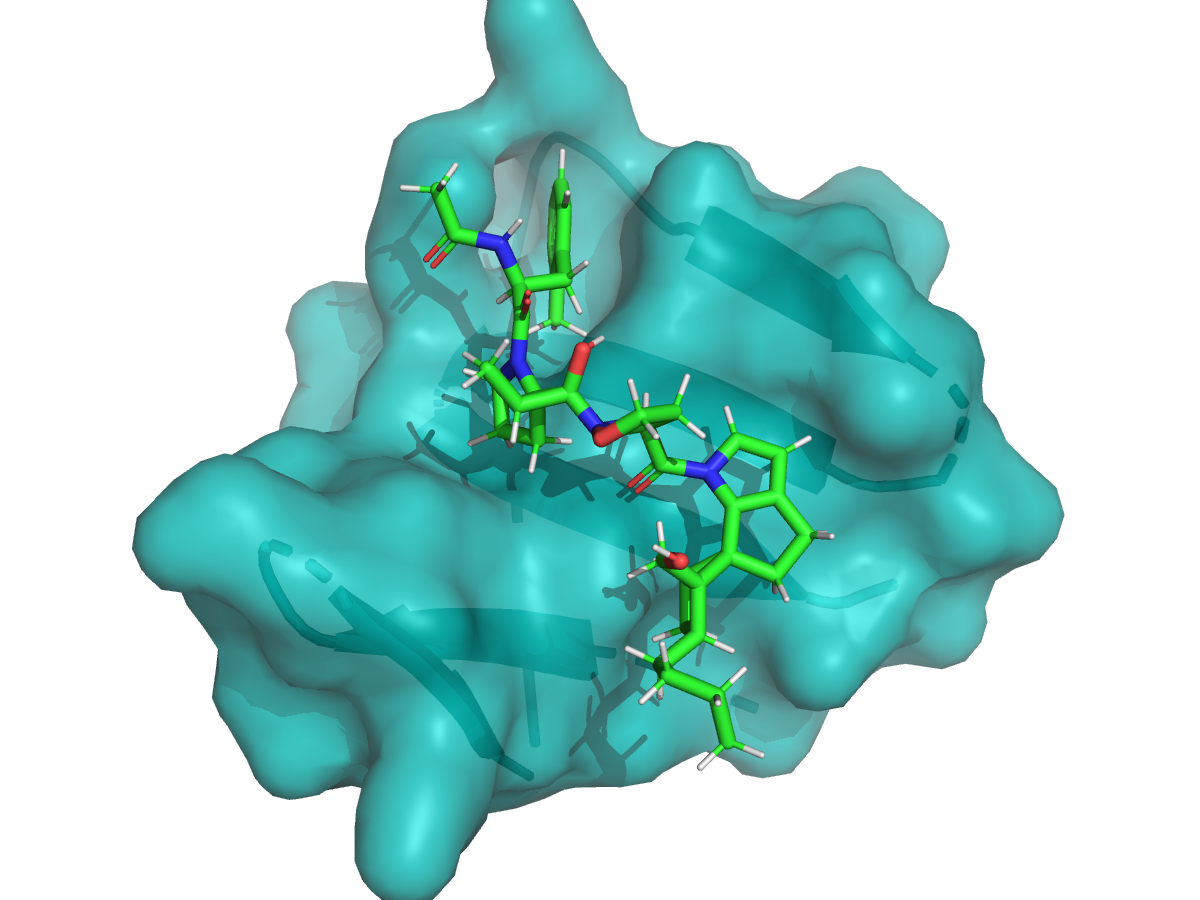} \hfill
    \includegraphics[width=0.3\textwidth]{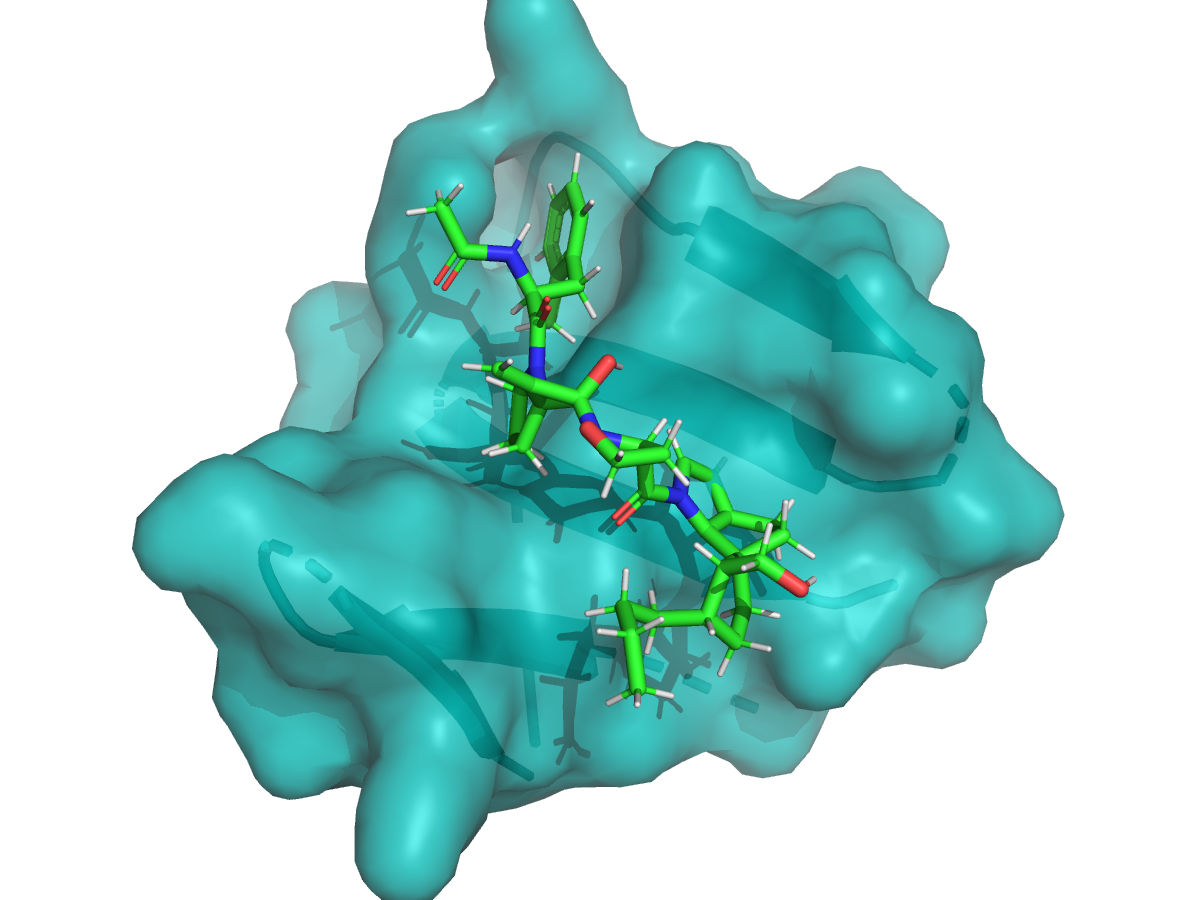}

    \caption{Nine molecules generated our IMG in single diffusion inference pass.}
    \label{fig:molecules}
\end{figure*}

In this section, we provide a qualitative visualization of molecules produced by our proposed Inference-time Multi-target Generation (IMG) algorithm. In this demonstration, IMG was employed with a batch size of $N=64$ and a resampling size of $M=32$, to optimize the multi-objectives: Vina score, SA score, and QED score. Figure~\ref{fig:molecules} displays 9 generated molecules by IMG with the target protein pocket (PDB code \texttt{5ndu}). These molecules are generated in a single diffusion inference pass, illustrating IMG's ability to generate a diverse set of high-quality molecules that optimized for multiple objectives.

\subsection{Ablation Study}

\begin{figure*}[ht]
    \centering
    \includegraphics[width=0.45\textwidth]{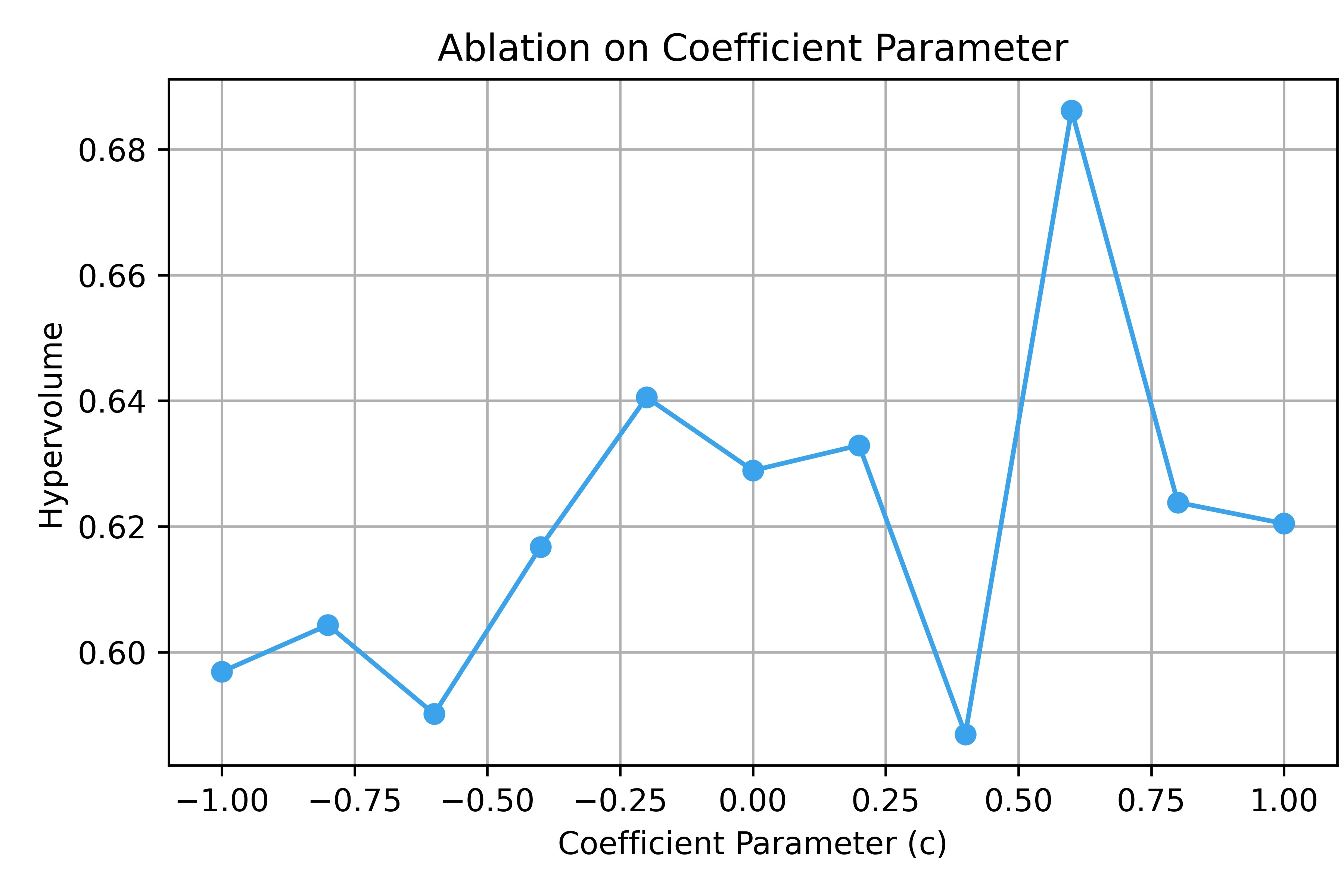} 
    \includegraphics[width=0.45\textwidth]{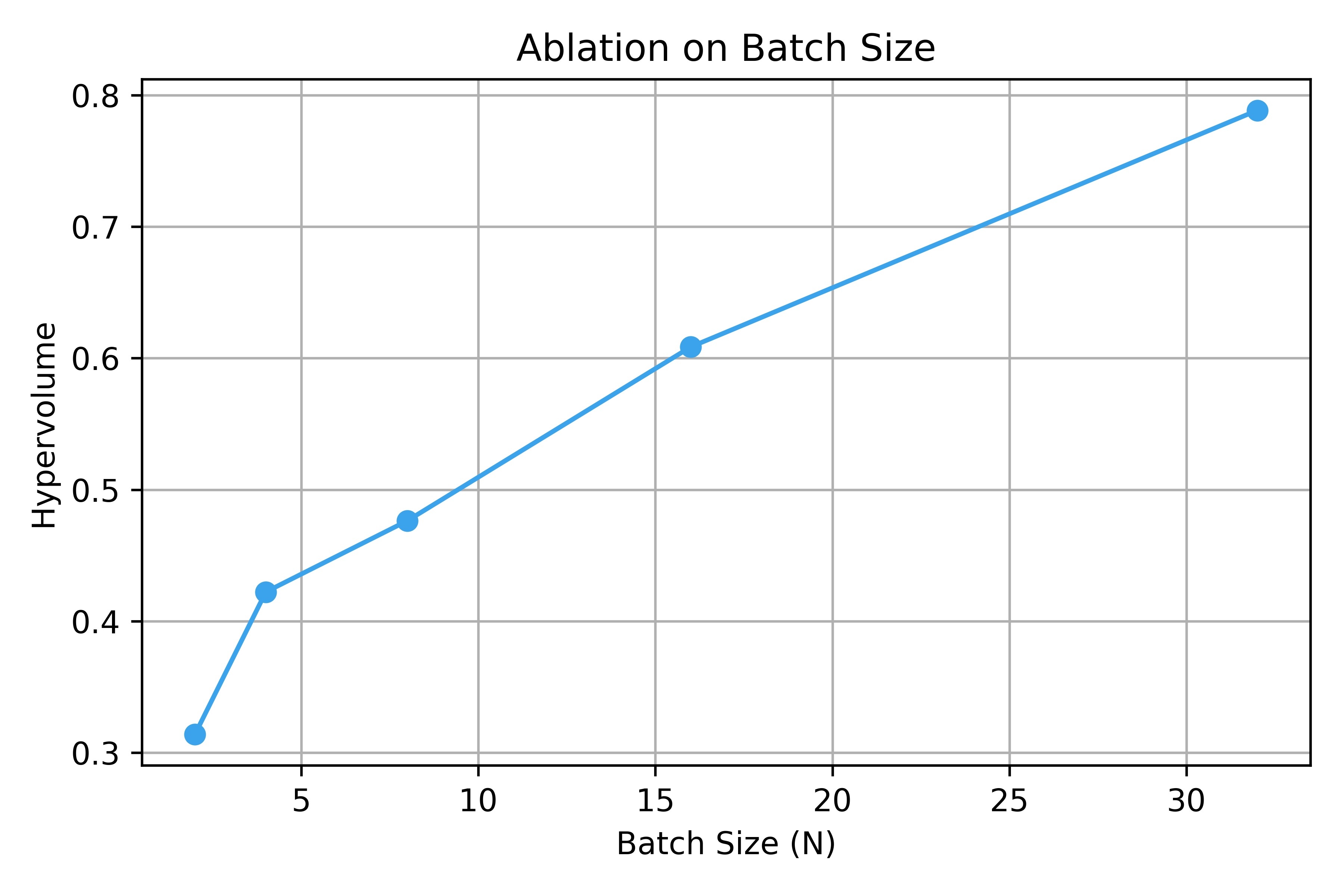} 
    \caption{An ablation study on the IMG algorithm's hyperparameters. \textbf{Left:} Hypervolume plotted against different values of the coefficient parameter ($c$), \textbf{Right:} Hypervolume plotted against different batch sizes ($N$).}
    \label{fig:ablation}
\end{figure*}

In this section, we perform ablation study to analyzes the sensitivity of our IMG algorithm's performance to two key hyperparameters: the coefficient parameter ($c$) and the batch size ($N$). The results are reported in Figure~\ref{fig:ablation}.

\textbf{Impact of the Coefficient Parameter.} The coefficient parameter $c_k$ is a component of the weighting function shown in Equation 10. Its optimal value depends on intractable normalization constants ($Z_k$), making it difficult to determine analytically. In our main experiments, we address this by dynamically setting each $c_k$ to the running upper bound for its corresponding $k$-th objective.

For this ablation study, we test a simplified, static approach. Instead of using individual, dynamic values, we set the coefficients for all $n$ objectives to be the same constant value, i.e., $c_1=c_2=\cdots=c_n=c$. We then investigate how performance, measured by hypervolume, is influenced as this uniform coefficient $c$ is varied. The experiment is performed with a fixed batch size of $N=32$ and a resampling size of $M=8$, while varying $c$ across the range $[-1.0,1.0]$. The results are reported in the left panel of Figure~\ref{fig:ablation}.

As shown in the figure, the hypervolume demonstrates robustness to the setting of $c$. While there are minor fluctuations, the overall performance is not highly sensitive to this parameter. This indicates that our IMG algorithm can achieve strong results without extensive hyperparameter tuning, highlighting its practicality and ease of use.

\textbf{Impact of the Batch Size.} The batch size $N$ not only determines the number of generated samples but also the number of distinct preference vectors ($\lambda^i$) in our IMG. It directly influences the diversity of the resulting solutions aimed at approximating the Pareto front. In this study, we investigated how performance, measured by hypervolume, scales with $N$. The experiment is performed with fixed resampling size of $M=32$, while varying the batch size $N\in \{2,4,8,16,32\}$. The results are reported in the right panel of Figure~\ref{fig:ablation}.

The figure shows a clear, positive correlation between the batch size and the final hypervolume. Increasing the batch size $N$ allows IMG to employ a wider set of distinct preference vectors ($\lambda^i$) simultaneously in a single pass. This parallel exploration of diverse trade-offs produces a more complete set of non-dominated solutions, thereby increasing the solution's hypervolume. Furthermore, increasing $N$ also expands the total candidate buffer ($B=N×M$) available at each resampling step, providing a larger pool from which to select optimal candidates for each preference vector.

\subsection{QMC Preference Vector}

\begin{figure*}
    \captionsetup[subfigure]{justification=centering}
    \centering

    \begin{subfigure}{0.24\textwidth}
        \includegraphics[width=\linewidth]{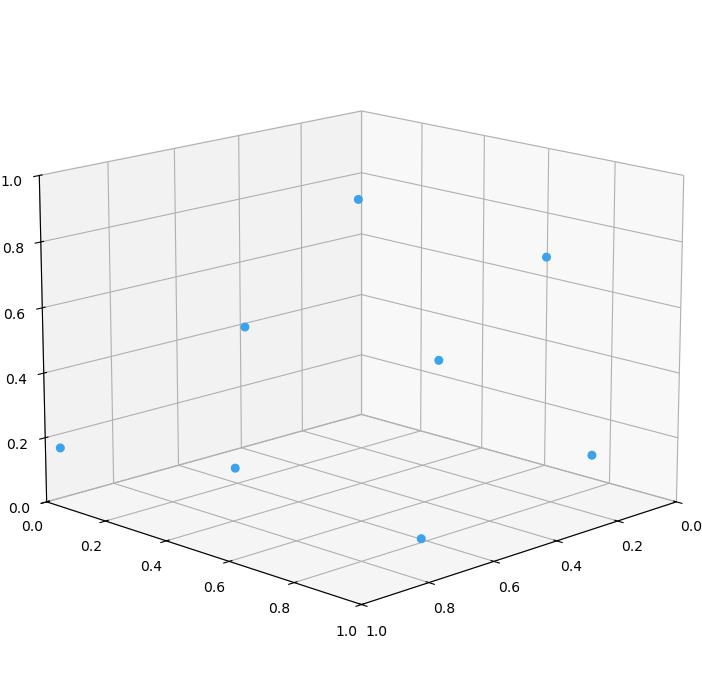}
        \caption*{{Our Alg 2\\($N=8$)}}
    \end{subfigure}\hfill
    \begin{subfigure}{0.24\textwidth}
        \includegraphics[width=\linewidth]{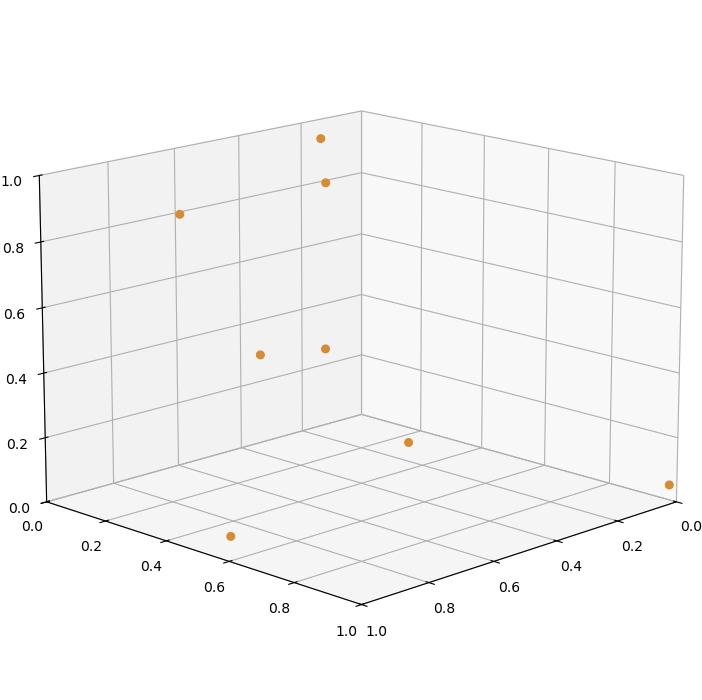}
        \caption*{{\cite{Spoints}\\($N=8$)}}
    \end{subfigure}\hfill
    \begin{subfigure}{0.24\textwidth}
        \includegraphics[width=\linewidth]{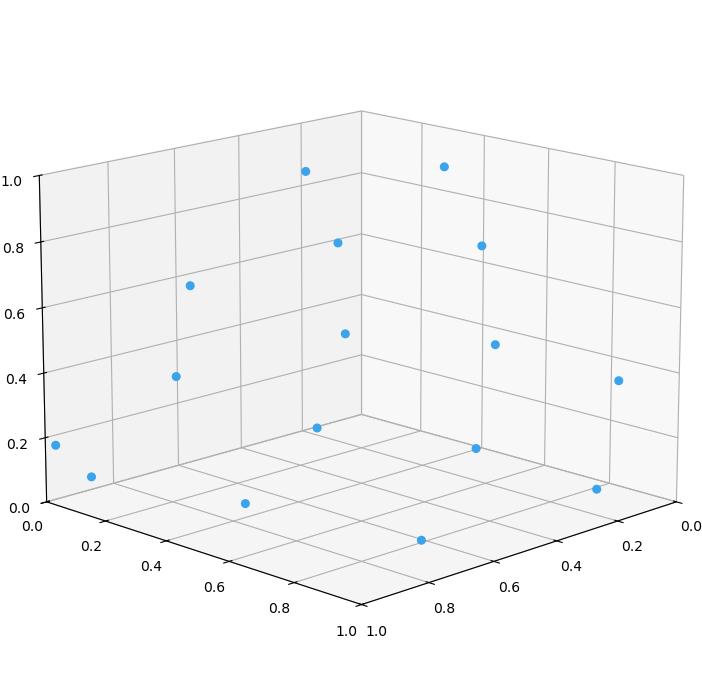}
        \caption*{{Our Alg 2\\($N=16$)}}
    \end{subfigure}\hfill
    \begin{subfigure}{0.24\textwidth}
        \includegraphics[width=\linewidth]{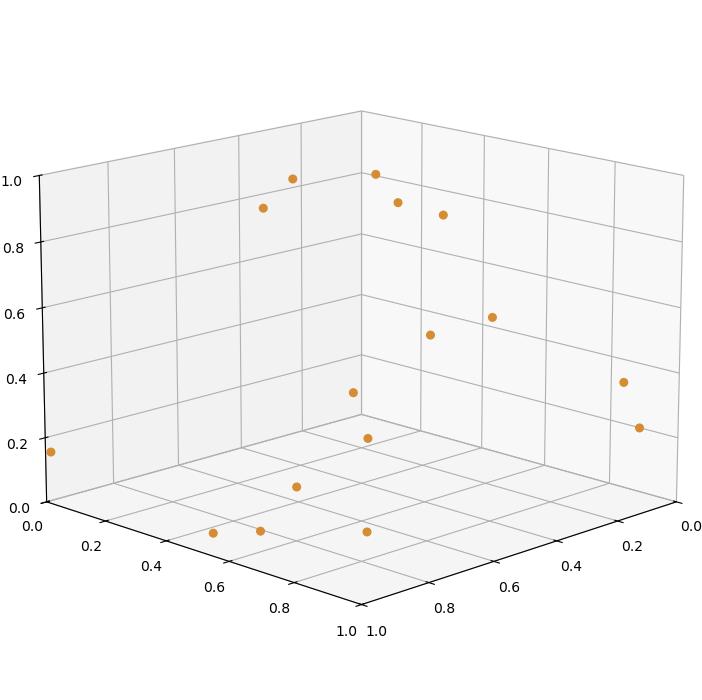}
        \caption*{{\cite{Spoints}\\($N=16$)}}
    \end{subfigure}

    \begin{subfigure}{0.24\textwidth}
        \includegraphics[width=\linewidth]{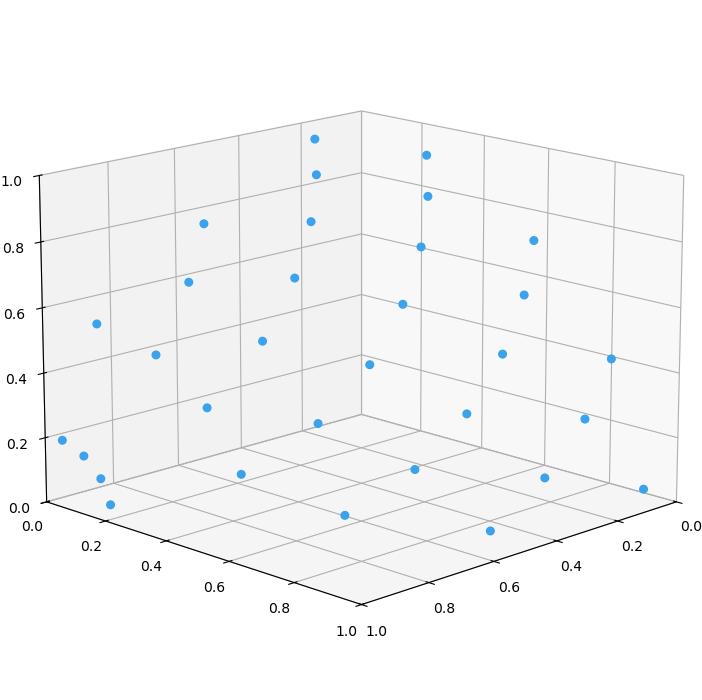}
        \caption*{{Our Alg 2\\($N=32$)}}
    \end{subfigure}\hfill
    \begin{subfigure}{0.24\textwidth}
        \includegraphics[width=\linewidth]{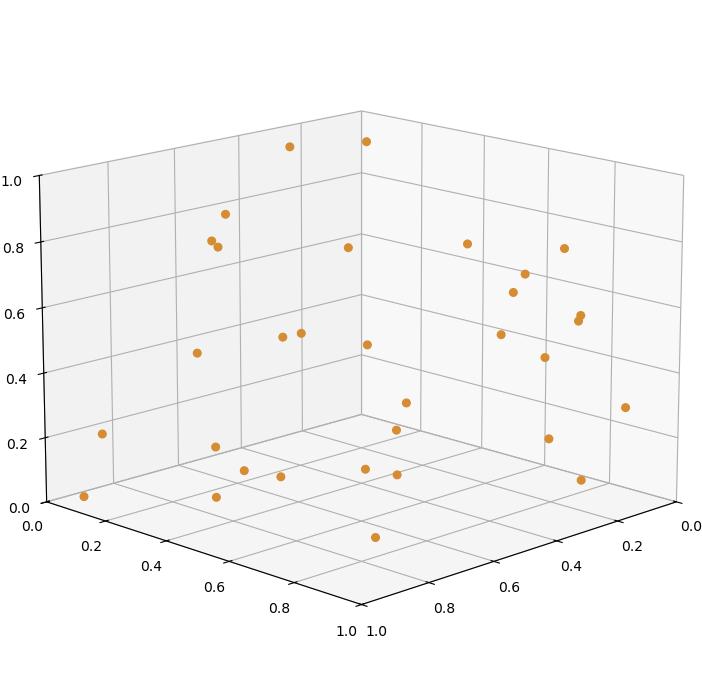}
        \caption*{{\cite{Spoints}\\($N=32$)}}
    \end{subfigure}\hfill
    \begin{subfigure}{0.24\textwidth}
        \includegraphics[width=\linewidth]{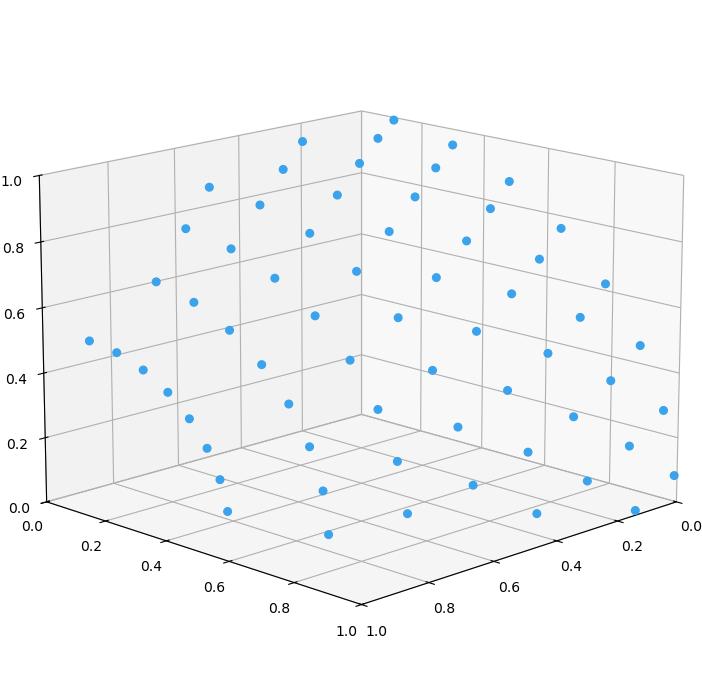}
        \caption*{{Our Alg 2\\($N=64$)}}
    \end{subfigure}\hfill
    \begin{subfigure}{0.24\textwidth}
        \includegraphics[width=\linewidth]{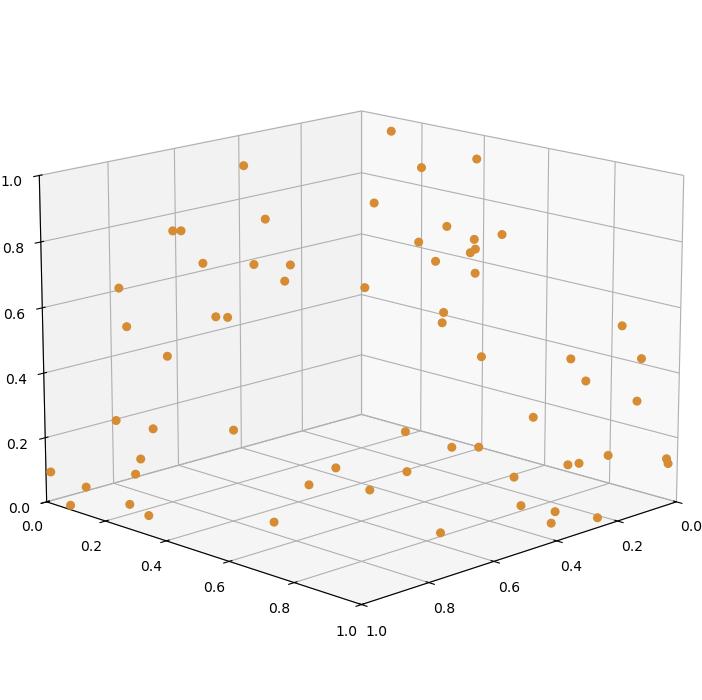}
        \caption*{{\cite{Spoints}\\($N=64$)}}
    \end{subfigure}

    \begin{subfigure}{0.24\textwidth}
        \includegraphics[width=\linewidth]{figures/lambda/qmc_N=100.jpeg}
        \caption*{{Our Alg 2\\($N=100$)}}
    \end{subfigure}\hfill
    \begin{subfigure}{0.24\textwidth}
        \includegraphics[width=\linewidth]{figures/lambda/iid_N=100.jpeg}
        \caption*{{\cite{Spoints}\\($N=100$)}}
    \end{subfigure}\hfill
    \begin{subfigure}{0.24\textwidth}
        \includegraphics[width=\linewidth]{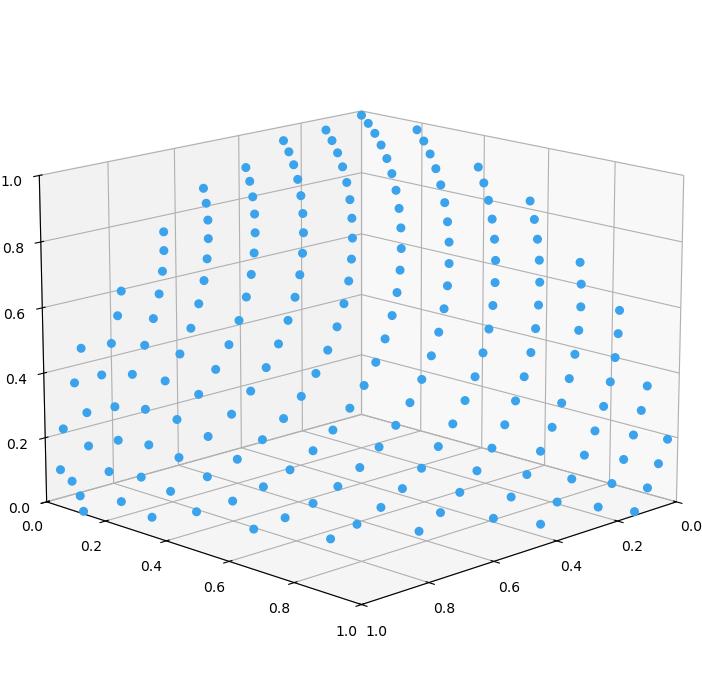}
        \caption*{{Our Alg 2\\($N=200$)}}
    \end{subfigure}\hfill
    \begin{subfigure}{0.24\textwidth}
        \includegraphics[width=\linewidth]{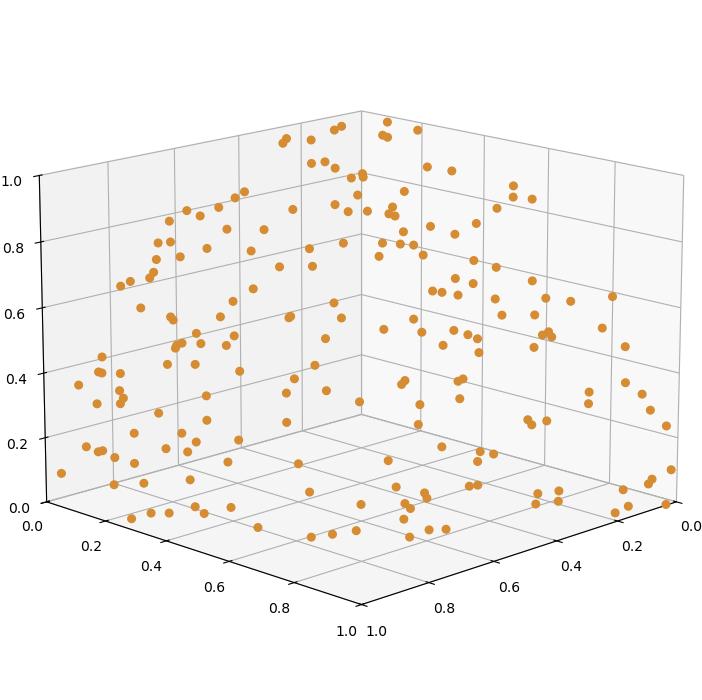}
        \caption*{{\cite{Spoints}\\($N=200$)}}
    \end{subfigure}

    \begin{subfigure}{0.24\textwidth}
        \includegraphics[width=\linewidth]{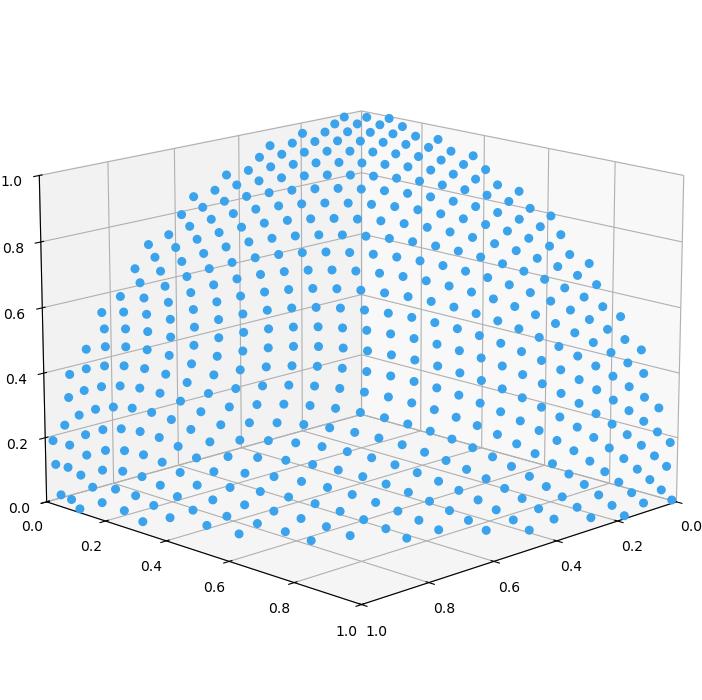}
        \caption*{{Our Alg 2\\($N=500$)}}
    \end{subfigure}\hfill
    \begin{subfigure}{0.24\textwidth}
        \includegraphics[width=\linewidth]{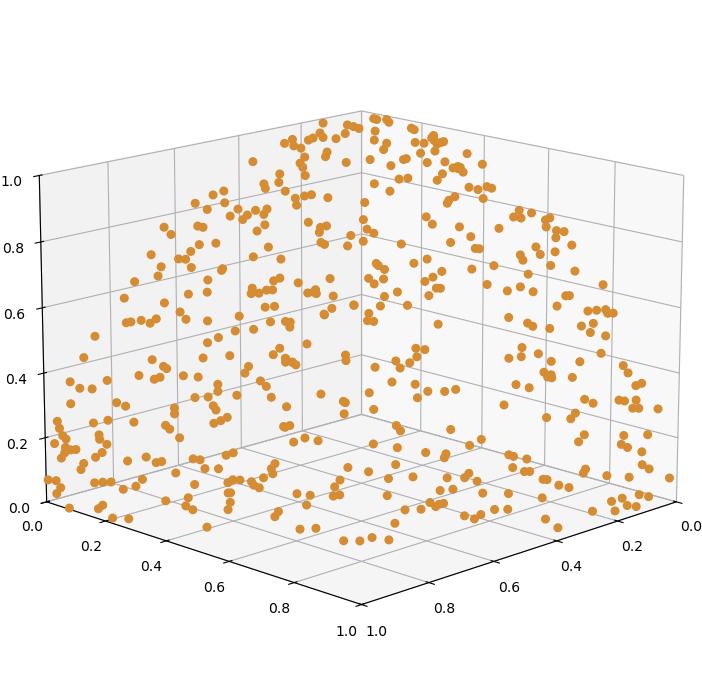}
        \caption*{{\cite{Spoints}\\($N=500$)}}
    \end{subfigure}\hfill
    \begin{subfigure}{0.24\textwidth}
        \includegraphics[width=\linewidth]{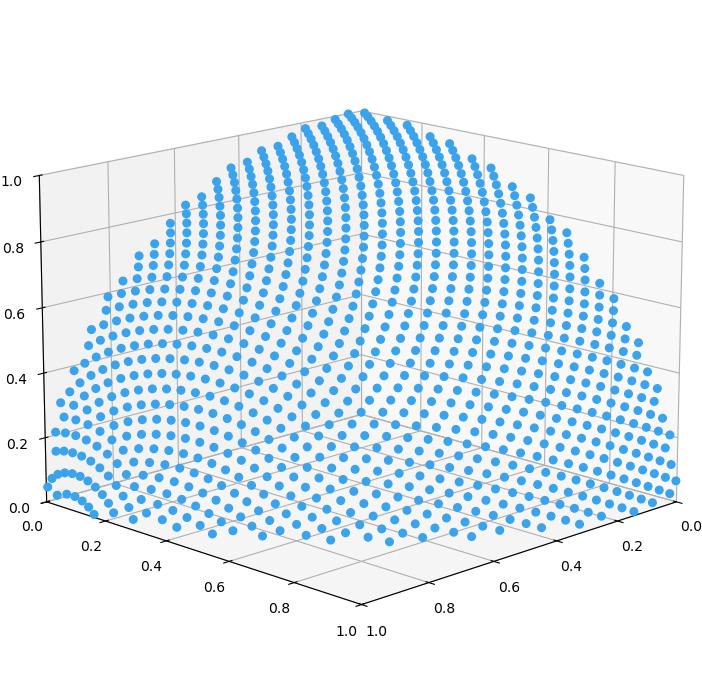}
        \caption*{{Our Alg 2\\($N=1000$)}}
    \end{subfigure}\hfill
    \begin{subfigure}{0.24\textwidth}
        \includegraphics[width=\linewidth]{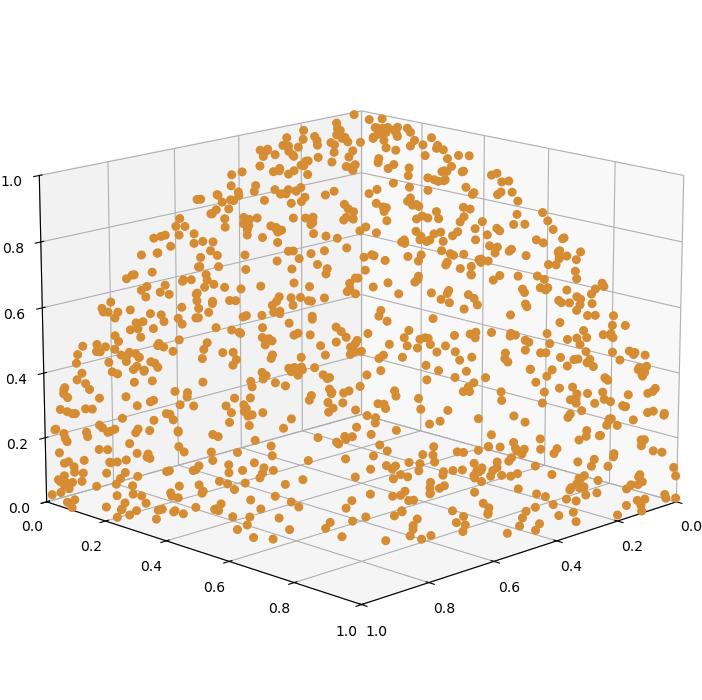}
        \caption*{{\cite{Spoints}\\($N=1000$)}}
    \end{subfigure}
    \caption{A visual comparison of preference vector generation methods for $n=3$ objectives across different sample sizes ($N$). For each sample size, we compare our proposed QMC-based method (Algorithm 2) against the uniform Monte Carlo sampling method from~\citet{Spoints}. Our method consistently produces more evenly distributed vectors, avoiding the clustering and gaps.}
    \label{fig:ablation_vectors}
\end{figure*}

In our main paper, we proposed Algorithm 2, a method based on Quasi-Monte Carlo (QMC) sampling, to generate evenly distributed preference vectors on the surface of the positive hyper-sphere. These vectors serve as an effective prior when user-specific preferences are not defined. Figure~\ref{fig:ablation_vectors} provides a side-by-side visual comparison of the vectors generated by our QMC-based algorithm against those produced by the uniform Monte Carlo sampling method from~\citet{Spoints}. The visualization demonstrates that our approach consistently avoids the clustering and sparse regions, resulting in more uniform coverage of the preference space.

\subsection{Solution's Pareto Front}

In this section, we visualizes the solutions' Pareto fronts in the 3D objective space. To ensure a fair comparison, all algorithms were configured to consume an equal $204.8$k objective function evaluations and produce $64$ solution samples. Specifically, our IMG algorithm used a batch size of $N=64$ and a resampling size of $M=32$, while baseline methods used a population size of $N=64$ and were optimized for $3200$ evolutionary steps.

\begin{figure*}[ht]
\centering
\includegraphics[width=0.5\linewidth]{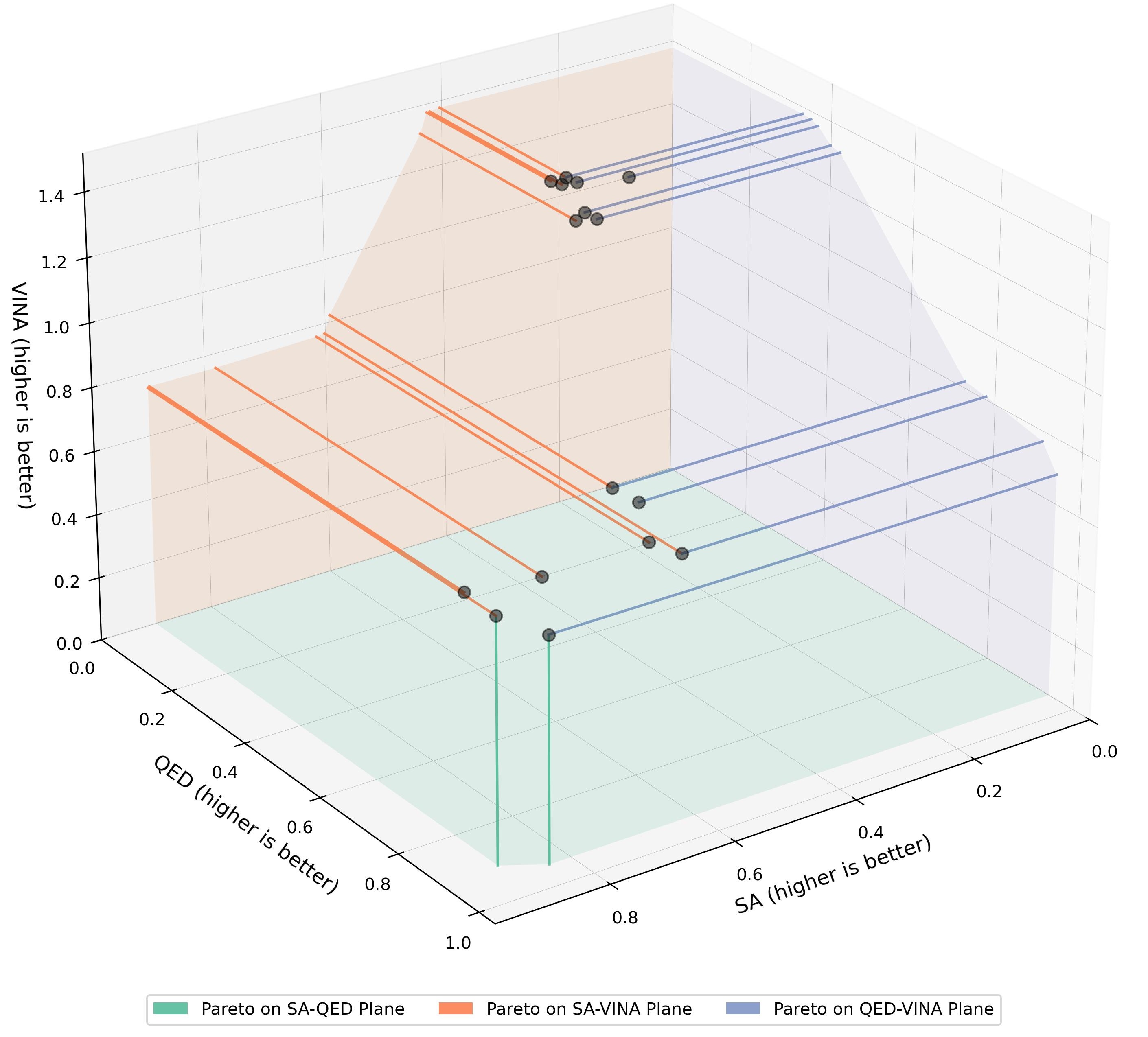}
\caption{The Pareto front generated from a single run of IMG. The plot displays the 16 non-dominated solutions identified from a set of 64 generated samples. Each axis represents one of the three objectives: Vina score, SA score, and QED score, where higher values indicate better performance. The colored projection line represents the edge of the Pareto front.}
\label{fig:pareto_img}
\end{figure*}

First, we analyze the solution set generated by IMG. From this solution set of $64$ samples, we identify 16 non-dominated Pareto points, which collectively achieve a hypervolume of $0.7640$. The resulting Pareto front is plotted in the 3D objective space in Figure~\ref{fig:pareto_img}.

\begin{figure*}[ht]
\centering
\includegraphics[width=0.5\linewidth]{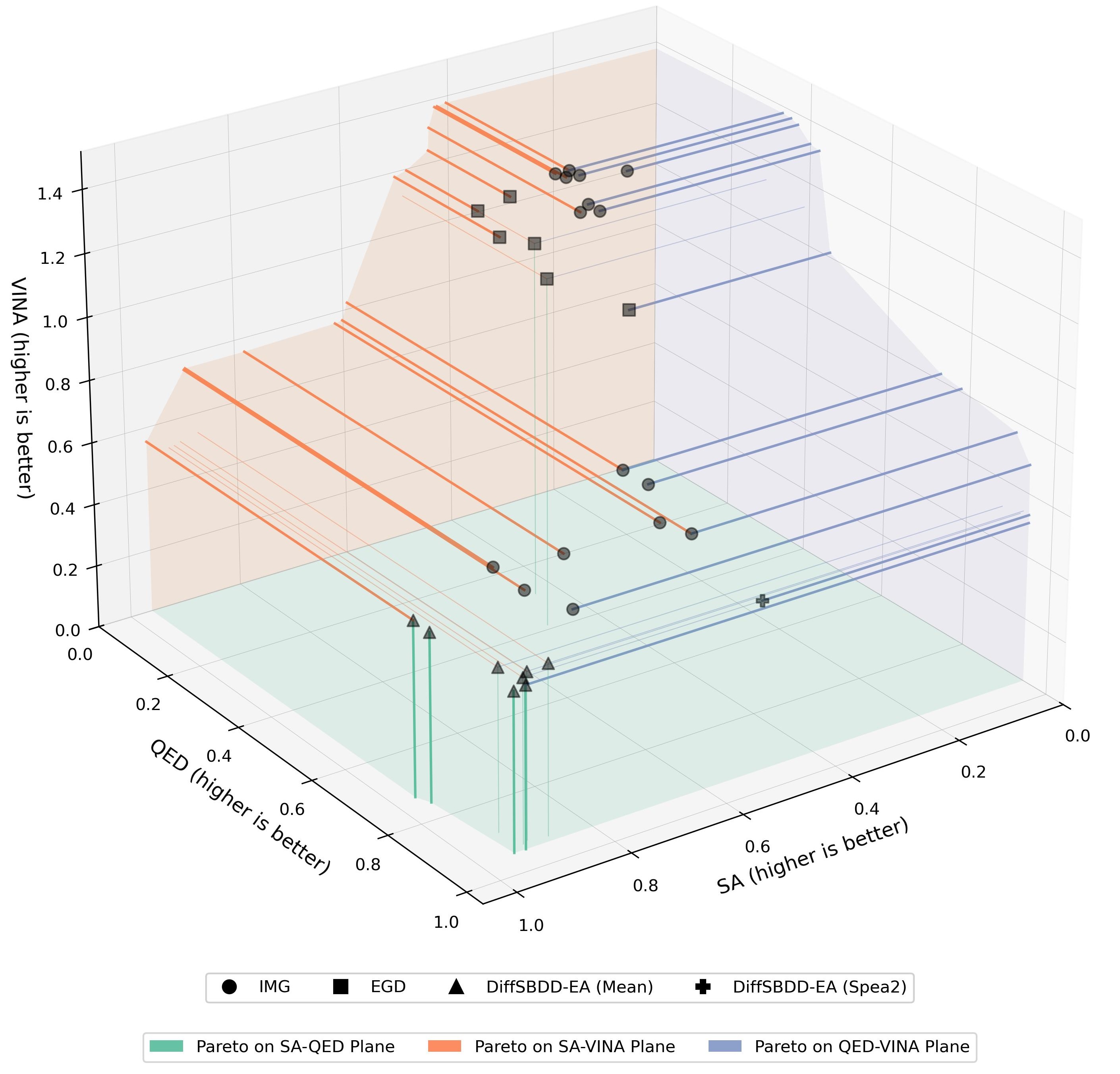}
\caption{The combined Pareto front from all four algorithms. The plot visualizes the $31$ non-dominated points from a total pool of $256$ combined solutions from all four algorithms. Different markers distinguish the solutions generated by each algorithm, illustrating that our IMG's solutions (circles) achieve a more diverse and even coverage of the objective space.}
\label{fig:pareto_relative}
\end{figure*}

Next, we perform a combined analysis by pooling the $64$ samples from each of the four algorithms: IMG, EGD, DiffSBDD-EA (Mean), and DiffSBDD-EA (Spea2), resulting in a total of 256 solutions. This combined set yields a combined Pareto front of $31$ non-dominated points with an overall hypervolume of $0.8103$. In Table~\ref{tab:pareto}, we report the contribution of each algorithm to this combined Pareto front. Notably, our IMG algorithm contributes 16 of these points, accounting for more than half of the entire Pareto front. This combined Pareto front is plotted in Figure~\ref{fig:pareto_relative}. We observe that IMG's solutions are distributed more evenly across the objective space, whereas the points from the baseline methods tend to be more concentrated.

\begin{table}[ht]
    \centering
    \caption{Number of Pareto points contributed by each algorithm to the combined Pareto front.}
    \label{tab:pareto}
    \begin{tabular}{|l|c|}
        \hline
        \textbf{Algorithm}   & \textbf{Number of Pareto Points} \\ \hline
        IMG                  & \textbf{16} \\ \hline
        EGD                  & 6 \\ \hline
        DiffSBDD-EA (Mean)      & 8 \\ \hline
        DiffSBDD-EA (Spea2)     & 1 \\ \hline
    \end{tabular}
\end{table}

\end{document}